\documentclass[11pt]{article}

\newcommand{\rUp}{\textcolor{green!45!black}{$\blacktriangle$}}   
\newcommand{\rDn}{\textcolor{red!70!black}{$\blacktriangledown$}}
\newcommand{\rMx}{\textcolor{orange!85!black}{$\blacklozenge$}}    
\newcommand{\rV}{\textcolor{blue!65!black}{$\mathbf{\vee}$}}       
\usepackage{tabularx}  
\newcolumntype{Y}{>{\raggedright\arraybackslash}X}
\usepackage{float}
\usepackage{listings}

\newcommand{\gcheck}{\textcolor{green!55!black}{\checkmark}}
\newcommand{\rcross}{\textcolor{red!70!black}{$\times$}}
\newcommand{\best}[1]{\textbf{#1}}
\newcommand{\second}[1]{\underline{#1}}

\usepackage{ragged2e}
\usepackage{xurl} % better line breaking for code-like strings

\newcommand{\fcode}[1]{\texttt{\footnotesize\path{#1}}}

\newcommand{\iterDot}[1]{%
  \ifcase#1\or
    $\bullet${\color{black!25}$\circ\circ$}\or
    {\color{black!25}$\circ$}$\bullet${\color{black!25}$\circ$}\or
    {\color{black!25}$\circ\circ$}$\bullet$\fi}

\usepackage[preprint]{acl}
\usepackage{hyperref}
\usepackage{amsmath}
\usepackage{amssymb}

\usepackage{booktabs}
\usepackage{multirow}
\usepackage{array}
\usepackage{threeparttable}
\usepackage{algorithm}
\usepackage{algorithmic}
\usepackage{siunitx}
\usepackage{adjustbox}
\usepackage[table]{xcolor}
\usepackage{enumitem}
\usepackage[most]{tcolorbox}
\usepackage{tikz}
\usetikzlibrary{shadows,shadows.blur}
\tcbuselibrary{skins,breakable,listings}
\definecolor{quantbg}{RGB}{248,249,246}
\definecolor{quantframe}{RGB}{20,48,44}
\definecolor{quanttag}{RGB}{120,88,43}
\definecolor{quantlabel}{RGB}{86,63,32}
\definecolor{quanttext}{RGB}{30,38,37}
\definecolor{quantcode}{RGB}{18,46,42}
\definecolor{quantshadow}{RGB}{218,224,218}

\newtcolorbox{quantprompt}[1]{
    enhanced,
    colback=quantbg,
    colframe=quantframe,
    boxrule=0.85pt,
    arc=5.5pt,
    left=6pt,
    right=6pt,
    top=6pt,
    bottom=6pt,
    before skip=5pt,
    after skip=5pt,
    width=\linewidth,
    fonttitle=\bfseries\scriptsize,
    coltitle=white,
    title={#1},
    drop shadow={
        shadow xshift=0.6pt,
        shadow yshift=-0.6pt,
        opacity=0.16,
        color=quantshadow
    },
    attach boxed title to top left={xshift=7pt,yshift=-2.4pt},
    boxed title style={
        enhanced,
        colback=quanttag,
        colframe=quanttag,
        boxrule=0pt,
        arc=4pt,
        left=6pt,
        right=6pt,
        top=1.1pt,
        bottom=1.1pt
    }
}

\newcommand{\promptlabel}[1]{{\bfseries\color{quantlabel}#1}}
\newcommand{\promptcontent}[1]{{\color{quanttext}#1}}
\newcommand{\promptcode}[1]{{\ttfamily\color{quantcode}#1}}
\definecolor{GIFTblue}{RGB}{165,202,248}
\definecolor{basebg}{RGB}{230,234,238}

\definecolor{win5}{RGB}{188,225,202}
\definecolor{win4}{RGB}{211,237,220}
\definecolor{win3}{RGB}{231,245,235}
\definecolor{win2}{RGB}{247,239,218}
\definecolor{win1}{RGB}{248,224,220}

\newcolumntype{N}{S[table-format=-2.2]}

\newcommand{\GIFThead}[1]{\cellcolor{GIFTblue}\textbf{#1}}
\newcommand{\basehead}[1]{\cellcolor{basebg}\textbf{#1}}
\newcommand{\winhead}[1]{\cellcolor{win3}\textbf{#1}}

\newcommand{\winner}[1]{\multicolumn{1}{N}{\bfseries #1}}

\newcommand{\winfive}[1]{\cellcolor{win5}{\bfseries #1}}
\newcommand{\winfour}[1]{\cellcolor{win4}{\bfseries #1}}
\newcommand{\winthree}[1]{\cellcolor{win3}{\bfseries #1}}
\newcommand{\wintwo}[1]{\cellcolor{win2}{\bfseries #1}}

\newcommand{\winzero}[1]{\cellcolor{win1}{\bfseries #1}}

\newcommand{\wincell}[1]{\cellcolor{win3}\textbf{#1}}

\usepackage{times}
\usepackage{latexsym}

\usepackage[T1]{fontenc}
\usepackage[utf8]{inputenc}

\usepackage{microtype}

\usepackage{inconsolata}

\usepackage{graphicx}

\title{GIFT: LLM-Guided State-Reward Interface for \\ Financial Reinforcement Learning}

\author{
\textbf{Yanyan Wu\textsuperscript{*1}},
\textbf{Boyi Zhang\textsuperscript{*2}},
\textbf{Yanlin Liu\textsuperscript{3}},
\textbf{Xinyu Fang\textsuperscript{2}},
\textbf{Jining Luan\textsuperscript{2}},
\textbf{Meiqi Zhang\textsuperscript{4}},
\\
\textbf{Jiacheng Liu\textsuperscript{2}},
\textbf{Hao Zeng\textsuperscript{5}},
\textbf{Dexu Yu\textsuperscript{6}},
\textbf{Chang Liu\textsuperscript{5}},
\textbf{Hanwen Du\textsuperscript{7}},
\textbf{Yongxin Ni\textsuperscript{8}},
\textbf{Youhua Li\textsuperscript{\textdagger 5}}
\\[0.55em]
{\footnotesize 
\textsuperscript{1}East China University of Science and Technology
\quad
\textsuperscript{2}University of Science and Technology of China}
\\[0.02em]
{\footnotesize 
\textsuperscript{3}Southwestern University of Finance and Economics
\quad
\textsuperscript{4}University of Sydney}
\\[0.02em]
{\footnotesize 
\textsuperscript{5}City University of Hong Kong
\quad
\textsuperscript{6}Northeastern University}
\\[0.2em]
{\footnotesize 
\textsuperscript{7}The Ohio State University
\quad
\textsuperscript{8}National University of Singapore}
\\[0.2em]
{\footnotesize
\texttt{\{23013032@mail.ecust.edu.cn, youhuali2-c@my.cityu.edu.hk\}}
}
\\[-0.1em]
{\footnotesize
\textsuperscript{*}Equal contribution.
\quad
\textsuperscript{\textdagger}Corresponding author.
}
}

\begin{document}
\maketitle

\begin{abstract}
Financial portfolio trading is naturally formulated as a reinforcement learning problem, where an agent sequentially rebalances assets under changing market conditions to balance return, risk, and transaction costs. 
Yet in non-stationary markets, raw OHLCV states and short-horizon return rewards often provide an under-specified learning interface, motivating large language models as a way to inject financial knowledge into state and reward design while constraining open-ended generation.
To this end, we propose GIFT (\textit{Guided Interface Design for Financial Trading}), an LLM-guided framework for state-reward interface design in PPO-based financial reinforcement learning. 
Rather than using the LLM to make trading decisions, GIFT uses Factor-guided State Enhancement (FSE) to generate state features from financial-factor primitives, Risk-rule-guided Reward Shaping (RRS) to generate auxiliary rewards from portfolio-risk rules, and Diagnostic-guided Refinement (DGR) to revise candidate interfaces using PPO rollout diagnostics. 
After refinement, GIFT fixes the selected state-reward interface before evaluation, with no further LLM queries or interface updates at test time.
Comprehensive rolling-window experiments across diverse market regimes and portfolio scenarios demonstrate that GIFT improves learning-signal quality and out-of-sample risk-adjusted portfolio performance over  baselines.
Code and data are available at: \url{https://github.com/KAG778/GIFT}.

\end{abstract}

\section{Introduction}

Financial trading (e.g., managing a portfolio of assets to maximize revenues) aims to dynamically adjust the weight of each asset over time to balance return, risk, and transaction costs~\cite{Wang_Huang_Tu_Zhang_Xu_2021}. Since each allocation decision depends on evolving market conditions and existing portfolio exposure, portfolio management is commonly formulated as a sequential decision-making problem~\cite{ijcai2020p641,10.24963/ijcai.2023/548,ijcai2022p557}. 
Deep reinforcement learning (DRL, e.g., PPO), as a dominant paradigm for sequential decision making, has therefore been widely used for portfolio trading and optimization.

In DRL-based portfolio trading, an agent learns rebalancing policies from historical market information and current portfolio holdings. 
The effectiveness of this learning process, however, depends critically on the state-reward interface provided to the agent~\cite{pmlr-v235-ma24l,pmlr-v235-wang24bh,10.5555/2772879.2772905}. 
On the state side, raw OHLCV observations (open, high, low, close prices and trading volume) mainly record basic price-volume movements.
Financial structures such as momentum, volatility, downside risk, liquidity, and mean reversion are often left implicit~\cite{10.1145/3637528.3672064}. 
The policy network must therefore recover these structures from noisy market sequences while simultaneously learning the trading policy, which makes state learning more difficult. 
On the reward side, short-horizon portfolio returns are affected by market shocks, transaction costs, and random fluctuations~\cite{ijcai2023p441}. 
Actions that reduce future risk may not receive immediate positive rewards, whereas actions with high short-term returns may increase drawdown or turnover. 
As a result, the reward signal can provide weak guidance for long-term return-risk objectives. 
Although prior studies have explored handcrafted state features, reward modification, and improved DRL algorithms, many still rely on fixed state and reward designs, making it difficult to adapt the learning interface across market windows~\cite{ijcai2020p623}.
\begin{figure*}[t]
\centering
\includegraphics[width=\textwidth]{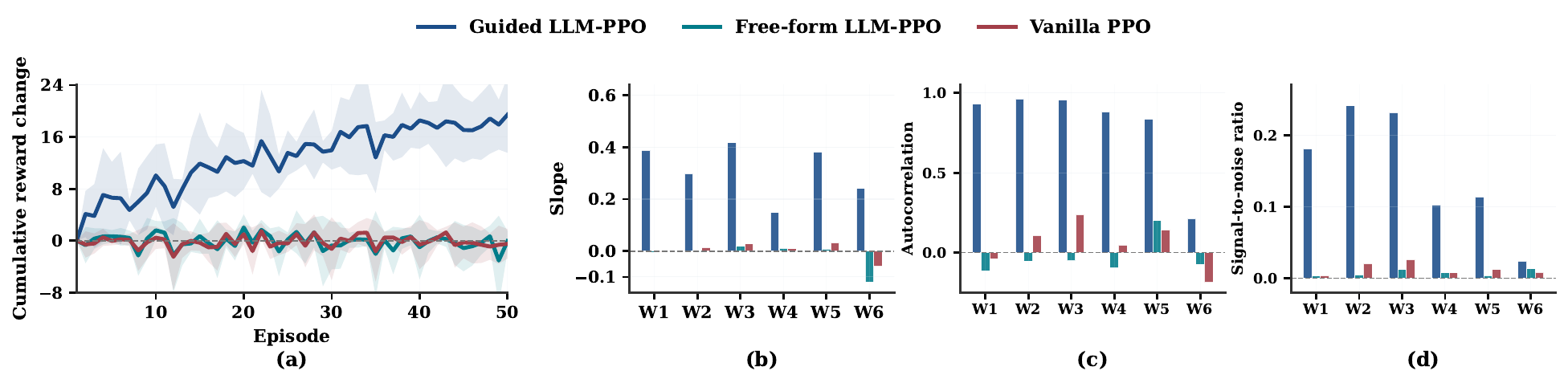}
\caption{
Diagnostic motivation experiment for PPO training feedback.
The figure compares three settings: Guided LLM-PPO, where LLM-generated state-reward designs are guided by financial knowledge; Free-form LLM-PPO, where LLM generation is not financially guided; and Vanilla PPO, which uses the original state and reward.
(a) shows the cumulative reward change during training.
(b)--(d) report slope, autocorrelation, and signal-to-noise ratio across rolling windows, measuring training improvement, reward-sequence consistency, and signal clarity, respectively.
Higher values indicate stronger progress, smoother reward dynamics, and cleaner learning signals.
The results suggest that Guided LLM-PPO provides more useful PPO training feedback than Free-form LLM-PPO and Vanilla PPO.
Details are provided in Appendix~\ref{app:preliminary_feedback}.
}
\label{fig:motivation}
\vspace{-2mm}
\end{figure*}

Large Language Models (LLMs) provide a way to inject financial knowledge into the state-reward interface of DRL. 
Given task descriptions, domain knowledge, and constraints, LLMs can generate executable feature and reward designs~\cite{Ma2023EurekaHR,10658032}. 
However, financial trading is not well suited to open-ended LLM generation. 
Because financial time series are noisy, non-stationary, and regime-dependent, unguided designs may follow short-term fluctuations, encode spurious correlations, or produce rewards that encourage excessive trading~\cite{10.1145/3580305.3599315}. 
This motivates a constrained use of LLMs: rather than using them to make trading decisions directly, we use them to design a more informative state-reward interface for the RL agent.

To examine whether such financial guidance can improve LLM-generated state-reward designs, we first conduct the diagnostic experiment in Figure~\ref{fig:motivation}.

\footnote{The definitions, calculation procedures, and window settings of these diagnostic metrics are provided in Appendix~\ref{app:preliminary_feedback}.}
The experiment compares financially guided LLM-PPO, Free-form LLM-PPO without financial-knowledge guidance, and Vanilla PPO. 

Guided LLM-PPO obtains higher reward improvement, reward-trend slope, reward autocorrelation, and signal-to-noise ratio in most windows.

To address the under-specified learning interface caused by raw states and short-horizon rewards, we propose GIFT (\textit{Guided Interface Design for Financial Trading}), an LLM-guided framework for state-reward interface design in PPO-based financial reinforcement learning. 
GIFT uses the LLM to generate executable state-enhancement functions and reward-shaping logic, while PPO remains responsible for learning the trading policy. 
Specifically, GIFT contains three components. 
1) Factor-guided State Enhancement (FSE) guides the LLM to generate additional state features from raw OHLCV inputs based on financial-factor primitives, such as momentum, volatility, downside risk, liquidity, and mean reversion. 
2) Risk-rule-guided Reward Shaping (RRS) guides the LLM to construct auxiliary rewards from portfolio-risk rules, including intrinsic reward terms, drawdown control, turnover penalty, volatility adjustment, and diversification.
3) Diagnostic-guided Refinement (DGR) evaluates each LLM-generated state-reward interface through PPO rollouts and feeds diagnostics on factor informativeness, reward stability, and portfolio behavior back to the LLM for subsequent revision.
After this offline interface design stage, GIFT fixes the selected interface before evaluation, with no further LLM queries or interface updates at test time.

The contributions of this paper are summarized as follows:
\begin{itemize}[leftmargin=*, itemsep=1pt, topsep=2pt, parsep=0.5pt, partopsep=0.5pt]
    \item To the best of our knowledge, we are the first to introduce an LLM-guided state-reward interface design perspective for financial reinforcement learning, where LLMs serve as constrained financial-knowledge-guided interface designers rather than direct trading agents.
    
    \item We propose GIFT, a PPO-based framework that combines Factor-guided State Enhancement, Risk-rule-guided Reward Shaping, and Diagnostic-guided Refinement to generate and revise executable state-reward interfaces.
    
    \item We conduct rolling-window experiments with variant, ablation, and robustness studies, demonstrating improved learning-signal quality and out-of-sample risk-adjusted portfolio performance.
\end{itemize}

\begin{figure*}[t]
\centering
\includegraphics[width=\textwidth]{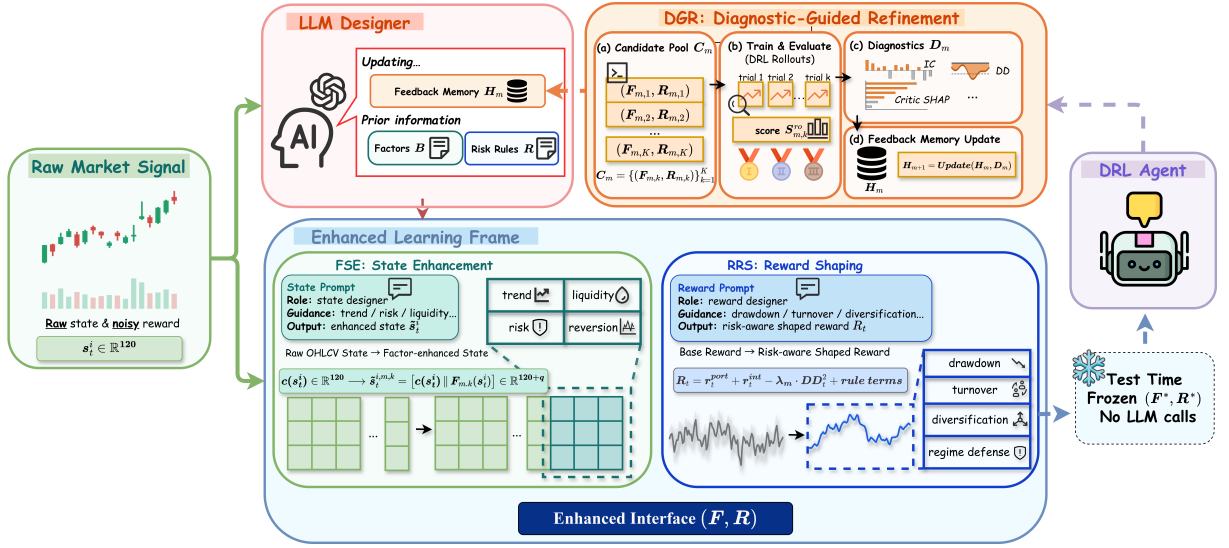}
\caption{
Overview of GIFT.
GIFT uses an LLM as a financial-knowledge-guided designer of the state-reward interface for PPO based portfolio trading.
Factor guided State Enhancement (FSE) augments raw OHLCV observations with compact financial factors, while Risk rule guided Reward Shaping (RRS) incorporates portfolio risk considerations into the training reward.
Diagnostic guided Refinement (DGR) evaluates candidate interfaces through PPO rollouts and uses the resulting diagnostics to guide the next generation round.
After LLM-guided generation and PPO based refinement, the selected interface \((F^\star,R^\star)\) is fixed for test time evaluation, with no further LLM calls or interface updates.
}
\label{fig:method}
\end{figure*}

\section{Method}
\label{sec:method}

\subsection{Problem Definition}
\label{sec:problem}

We formulate portfolio trading as a finite-horizon sequential decision-making problem.
At each trading day \(t\), the agent observes market signals and portfolio context, and chooses portfolio weights for the next holding period.
A trajectory is denoted as
\begin{equation}
\tau=\{(s_t,a_t,r_t)\}_{t=0}^{T-1},
\label{eq:trading_trajectory}
\end{equation}
where \(s_t\in\mathcal{S}\) is the state, \(a_t\in\mathcal{A}\) is the allocation action, and \(r_t\) is the portfolio reward.

The action is a long-only allocation over \(N\) risky assets and one cash position:
\begin{equation}
a_t=(w_{t,1},\ldots,w_{t,N},w_{t,\mathrm{cash}}),
\label{eq:action_space}
\end{equation}
where \(w_{t,i}\ge0\), \(w_{t,\mathrm{cash}}\ge0\), and
\(\sum_{i=1}^{N}w_{t,i}+w_{t,\mathrm{cash}}=1\).

Let \(y_t\) denote the asset-return vector over the next holding period and \(c\) the proportional transaction-cost rate.
The base portfolio reward is
\begin{equation}
r_t^{\mathrm{port}}
=
a_t^\top y_t
-
c \|a_t-a_{t-1}\|_1 ,
\label{eq:base_reward}
\end{equation}
where the second term penalizes turnover.
The goal is to learn a policy \(\pi_\theta(a_t\mid H_t)\) that maximizes expected portfolio reward using only information available before action selection.

\subsection{Overview of GIFT}
\label{sec:overview}

Figure~\ref{fig:method} illustrates GIFT.
GIFT uses the LLM \(M\) as a constrained designer of the state-reward interface for PPO, rather than as a trading agent.
It has three components: \textit{Factor-guided State Enhancement} (FSE), which generates executable factor-based state enhancements from OHLCV inputs; \textit{Risk-rule-guided Reward Shaping} (RRS), which constructs auxiliary rewards from portfolio-risk rules; and \textit{Diagnostic-guided Refinement} (DGR), which evaluates candidate interfaces through PPO rollouts and uses diagnostics to guide revision.
After offline interface design, the selected interface is fixed for evaluation.

\subsection{Factor-guided State Enhancement}
\label{sec:fse}

Raw OHLCV histories contain useful trading signals, but many financial structures are not explicit in the input representation.
A policy trained directly on these sequences must infer trend, volatility, downside risk, liquidity, and mean-reversion patterns while also learning allocation behavior.
This entangles representation learning with policy optimization and can make training unstable across market regimes.

FSE constrains LLM generation with a financial factor library \(\mathcal{B}\).
In this work, \(\mathcal{B}\) contains factor primitives for momentum, volatility, downside risk, liquidity, mean reversion, and trend strength.
Instead of allowing arbitrary feature construction, FSE asks the LLM to select, transform, and compose a compact set of factor channels.

At refinement round \(m\), the LLM generates \(K\) candidate state-enhancement functions:
\begin{equation}
\mathcal{F}_m=\{F_{m,k}\}_{k=1}^{K},
\qquad
F_{m,k}\sim M(d_m;\mathcal{B}),
\label{eq:fse_generation}
\end{equation}
where \(d_m\) is the structured prompt containing market statistics and feedback from previous rounds.
For asset \(i\), a candidate function augments the compact raw-market encoding as
\begin{equation}
\tilde{s}_{t}^{i,m,k}
=
\left[
c(s_t^i)
\;\Vert\;
F_{m,k}(s_t^i)
\right],
\label{eq:enhanced_state}
\end{equation}
where \(c(\cdot)\) denotes the compact price/return encoding and \(F_{m,k}(\cdot)\) denotes the generated factor transformation.
The enhanced asset representations are then combined with portfolio context to form the PPO observation.

\subsection{Risk-rule-guided Reward Shaping}
\label{sec:rrs}

Short-horizon portfolio return is a natural reward, but it is often noisy and weakly aligned with long-term return-risk objectives.
For example, a policy may obtain temporary return gains while increasing drawdown, turnover, concentration, or exposure to high-volatility regimes.
RRS addresses this issue by augmenting the base return with financial risk-control signals.

Given the enhanced state from FSE, the LLM generates an intrinsic reward term and selects a small subset of risk rules from a library \(\mathcal{R}\).
The library contains portfolio-level rules for diversification and concentration control, turnover penalty, drawdown control, volatility scaling, regime defense, and momentum alignment.
For readability, we omit the candidate index \(k\) in the following formulation.
The shaped reward is
\begin{equation}
R_t
=
r_t^{\mathrm{port}}
-
\lambda_m \mathrm{DD}_t^2
+
r^{\mathrm{LLM}}_{\mathrm{int},t}
+
\sum_{j\in\mathcal{S}^{R}_m}
\alpha_{m,j}\rho_j(\tau_t),
\label{eq:rrs_reward}
\end{equation}
where \(\mathrm{DD}_t\) is the current drawdown, \(r^{\mathrm{LLM}}_{\mathrm{int},t}\) is the LLM-generated intrinsic reward computed from the enhanced state, and \(\rho_j(\tau_t)\) denotes a behavior-specific risk rule.
The drawdown coefficient \(\lambda_m\), selected rule subset \(\mathcal{S}^{R}_m\), and rule weights \(\alpha_{m,j}\) are determined during the offline interface design stage.

RRS does not replace the portfolio-return objective.
Instead, it supplies auxiliary feedback that makes the training reward more sensitive to risk and allocation behavior that one-step return alone may fail to distinguish.

\subsection{Diagnostic-guided Refinement}
\label{sec:feedback}

A single LLM-generated interface may include ineffective factors or poorly scaled reward terms.
DGR therefore closes the loop between generated interfaces and PPO training outcomes.
For each candidate interface \((F_{m,k},R_{m,k})\), GIFT trains PPO on the historical design window and computes diagnostics from the resulting rollouts.

The diagnostics summarize three aspects of the candidate interface.
State diagnostics measure factor informativeness using channel IC, regime-specific IC, and critic attribution signals.
Reward diagnostics examine whether the shaped reward provides stable learning feedback, using reward trend and reward variability.
Portfolio diagnostics describe the induced trading behavior, including return, drawdown, and risk-adjusted performance.
Together, these diagnostics identify failure modes such as weak factors, unstable intrinsic rewards, or risk rules that do not effectively control drawdown.

The diagnostic summary \(\mathcal{D}_m\) is compressed into a feedback memory:
\begin{equation}
\mathcal{H}_{m+1}
=
\mathrm{Update}(\mathcal{H}_m,\mathcal{D}_m).
\label{eq:feedback_update}
\end{equation}
The next prompt uses \(\mathcal{H}_{m+1}\) as revision guidance, for example to remove low-IC factors, rescale unstable reward terms, or strengthen risk-control rules under high drawdown.
The next-round candidates are generated as
\begin{equation}
(F_{m+1,k},R_{m+1,k})
\sim
M(\mathcal{B},\mathcal{R},\mathrm{Stats}_{m+1},\mathcal{H}_{m+1}).
\label{eq:dgr_generation}
\end{equation}

After \(M\) refinement rounds, GIFT selects the best executable interface according to rollout performance on the historical design window and freezes it as \((F^\star,R^\star)\).
The resulting selection patterns, including selected factors, selected risk rules, intrinsic-reward structures, and selected refinement rounds, are analyzed in Section~\ref{sec:semantic_analysis}.

\subsection{Offline Interface Design and Evaluation}
\label{sec:selection}

GIFT separates offline interface design from policy evaluation.
For each rolling window, the LLM uses only historical data before the evaluation period to generate and refine candidate interfaces.
Each executable candidate is trained with PPO and ranked by a rollout-level score on the historical design window.
The scoring rule and validation checks are provided in Appendix~\ref{app:rollout_scoring}.

After offline design, the selected interface \((F^\star,R^\star)\) is fixed.
During evaluation, this frozen interface is attached to a newly initialized PPO agent.
The agent follows the same chronological policy-adaptation protocol as the Pure PPO baseline: it adapts on the first half of the test window and is evaluated on the second half.
Only the interface is frozen; the policy-learning protocol is unchanged across methods.
No LLM query, prompt update, feedback-memory write, or interface modification is allowed during evaluation.
This protocol ensures that GIFT transfers a selected state-reward interface to the evaluation window, rather than adapting the interface using future information.

\section{Experiments}

\begin{itemize}[leftmargin=*, itemsep=0pt, topsep=1pt, parsep=0pt, partopsep=0pt]
    \item \textbf{RQ1:} Does GIFT improve out-of-sample portfolio performance across rolling market windows and sector portfolios?
    \item \textbf{RQ2:} Do financially guided state-reward interfaces provide better PPO training feedback than fixed or free-form designs?
    \item \textbf{RQ3:} Which components of GIFT contribute to the improvement?
    \item \textbf{RQ4:} How does GIFT select different state-reward interfaces across rolling market windows?
    \item \textbf{RQ5:} Is the selected state-reward interface transferable and robust under different RL backbones, offline design settings, random seeds, and baseline strengths?
\end{itemize}

\subsection{Experimental Setup}
\label{sec:exp_setting}

\paragraph{Data and environment.}
We evaluate daily portfolio allocation on S\&P 500 constituents over six rolling test windows, summarized in Table~\ref{tab:market_windows}.
The experiments use six portfolio panels: Technology, Healthcare, Energy, Industrials and two mixed-sector panels, Light Mix and Heavy Mix.
Each panel contains five representative stocks and cash; the detailed stock composition is reported in the Appendix~\ref{app:portfolio_panels}.
We use Light Mix for detailed behavioral analysis.
The observation contains 20 trading days of OHLCV-related features, the action is a long-only allocation over the five stocks and cash, and the transaction cost is \(0.1\%\) per unit turnover.

\begin{table}[t]
\centering
\footnotesize
\setlength{\tabcolsep}{4pt}
\renewcommand{\arraystretch}{1.05}                  
\caption{Rolling test windows and market regimes (2020--2023).
\rUp\,bull, \rDn\,bear, \rMx\,mixed, \rV\,V-shaped.}
\label{tab:market_windows}
\begin{tabularx}{\columnwidth}{@{}c l c >{\scriptsize\raggedright\arraybackslash}X@{}}
\toprule
\textbf{ID} & \textbf{Test period} & \textbf{Type} & \textbf{Regime} \\
\midrule
W1 & Dec'20--Jun'21 & \rUp & Post-pandemic recovery bull \\
W2 & Jul'21--Dec'21 & \rUp & Pre-tightening bull \\
W3 & Jan'22--Jun'22 & \rDn & Tightening and geopolitical-shock bear \\
W4 & Jul'22--Dec'22 & \rMx & High-rate bottoming \\
W5 & Jan'23--Jun'23 & \rMx & Narrow bull with banking stress and AI boom \\
W6 & Jul'23--Dec'23 & \rV  & V-shaped recovery after rate shock \\
\bottomrule
\end{tabularx}
\end{table}

\paragraph{Baseline and evaluation protocol.}
Since our focus is the proposed interface-design paradigm, the main experiments compare GIFT-enhanced PPO with a Pure PPO baseline under the same policy architecture, hyperparameters, and chronological split.
This controlled comparison isolates the effect of the LLM-guided state-reward interface.
Additional comparisons with traditional trading baselines are provided in Appendix Table~\ref{tab:gift_vs_trading_baselines}.

For GIFT, offline interface design uses only data before the current evaluation window, and the selected state-enhancement and reward-shaping code is fixed before evaluation.
During evaluation, no LLM query, prompt update, feedback-memory write, or interface modification is allowed.
Both GIFT and Pure PPO follow the same policy-adaptation protocol: the agent adapts on the first half of each test window and is evaluated on the second half. 

\vspace{-2mm}

\paragraph{Metrics.}
We report five portfolio metrics.
Cumulative return measures total profit over the evaluation period.
Sharpe ratio measures return per unit of volatility, while Sortino ratio focuses on downside volatility.
Maximum drawdown (MDD) measures the largest peak-to-trough loss.
Calmar ratio compares return against maximum drawdown.
Cumulative return, Sharpe, Sortino, and Calmar are better when higher, whereas MDD is better when lower.
\begin{table*}[!t]
\centering
\caption{
Out-of-sample performance of GIFT against the Pure PPO baseline under the same architecture, hyperparameters, and chronological split.
Return and MDD are reported in percentages.
}
\label{tab:main_performance_all_sectors}

\scriptsize
\setlength{\tabcolsep}{2.55pt}
\renewcommand{\arraystretch}{1.08}

\begin{adjustbox}{width=\textwidth}
\begin{tabular}{l *{2}{*{10}{N} c}}
\toprule
%\multicolumn{23}{c}{\textbf{GIFT vs. Pure PPO}} \\
%\midrule

% ====================== Panel A: Technology / Healthcare ======================
\textbf{Window}
& \multicolumn{11}{c}{\textbf{Technology}}
& \multicolumn{11}{c}{\textbf{Healthcare}} \\
\cmidrule(lr){2-12}
\cmidrule(lr){13-23}
& \multicolumn{2}{c}{\textbf{Ret. (\%) $\uparrow$}}
& \multicolumn{2}{c}{\textbf{Sharpe $\uparrow$}}
& \multicolumn{2}{c}{\textbf{Sortino $\uparrow$}}
& \multicolumn{2}{c}{\textbf{MDD (\%) $\downarrow$}}
& \multicolumn{2}{c}{\textbf{Calmar $\uparrow$}}
& \winhead{Win}
& \multicolumn{2}{c}{\textbf{Ret. (\%) $\uparrow$}}
& \multicolumn{2}{c}{\textbf{Sharpe $\uparrow$}}
& \multicolumn{2}{c}{\textbf{Sortino $\uparrow$}}
& \multicolumn{2}{c}{\textbf{MDD (\%) $\downarrow$}}
& \multicolumn{2}{c}{\textbf{Calmar $\uparrow$}}
& \winhead{Win} \\
\cmidrule(lr){2-3}
\cmidrule(lr){4-5}
\cmidrule(lr){6-7}
\cmidrule(lr){8-9}
\cmidrule(lr){10-11}
\cmidrule(lr){13-14}
\cmidrule(lr){15-16}
\cmidrule(lr){17-18}
\cmidrule(lr){19-20}
\cmidrule(lr){21-22}
& \multicolumn{1}{c}{\GIFThead{GIFT}} & \multicolumn{1}{c}{\basehead{Pure PPO}}
& \multicolumn{1}{c}{\GIFThead{GIFT}} & \multicolumn{1}{c}{\basehead{Pure PPO}}
& \multicolumn{1}{c}{\GIFThead{GIFT}} & \multicolumn{1}{c}{\basehead{Pure PPO}}
& \multicolumn{1}{c}{\GIFThead{GIFT}} & \multicolumn{1}{c}{\basehead{Pure PPO}}
& \multicolumn{1}{c}{\GIFThead{GIFT}} & \multicolumn{1}{c}{\basehead{Pure PPO}}
& \winhead{}
& \multicolumn{1}{c}{\GIFThead{GIFT}} & \multicolumn{1}{c}{\basehead{Pure PPO}}
& \multicolumn{1}{c}{\GIFThead{GIFT}} & \multicolumn{1}{c}{\basehead{Pure PPO}}
& \multicolumn{1}{c}{\GIFThead{GIFT}} & \multicolumn{1}{c}{\basehead{Pure PPO}}
& \multicolumn{1}{c}{\GIFThead{GIFT}} & \multicolumn{1}{c}{\basehead{Pure PPO}}
& \multicolumn{1}{c}{\GIFThead{GIFT}} & \multicolumn{1}{c}{\basehead{Pure PPO}}
& \winhead{} \\
\midrule

W1
& \winner{21.61} & 20.07
& \winner{2.40} & 2.27
& \winner{2.39} & 2.14
& \winner{10.28} & 10.41
& \winner{4.82} & 4.46
& \winfive{5/5}
& 10.54 & \winner{10.78}
& \winner{2.39} & 2.24
& \winner{2.77} & 2.59
& \winner{2.65} & 3.29
& \winner{9.37} & 7.74
& \winfour{4/5} \\

W2
& \winner{11.15} & 8.16
& \winner{1.64} & 1.03
& \winner{1.58} & 0.99
& \winner{7.86} & 10.42
& \winner{3.43} & 2.03
& \winfive{5/5}
& \winner{16.20} & 7.32
& \winner{2.38} & 1.22
& \winner{2.89} & 1.34
& \winner{9.74} & 11.65
& \winner{3.86} & 1.57
& \winfive{5/5} \\

W3
& \winner{-22.01} & -25.41
& \winner{-1.50} & -1.74
& \winner{-1.43} & -1.67
& \winner{25.48} & 28.26
& \winner{-2.11} & -2.27
& \winfive{5/5}
& \winner{12.45} & 11.27
& \winner{1.68} & 1.49
& \winner{1.96} & 1.67
& \winner{8.09} & 8.54
& \winner{3.71} & 3.23
& \winfive{5/5} \\

W4
& \winner{-13.86} & -17.37
& \winner{-0.99} & -1.17
& \winner{-1.12} & -1.29
& \winner{24.06} & 26.82
& \winner{-1.30} & -1.50
& \winfive{5/5}
& \winner{6.41} & 5.12
& \winner{1.17} & 0.88
& \winner{1.22} & 0.86
& 6.23 & \winner{5.49}
& \winner{2.57} & 2.41
& \winfour{4/5} \\

W5
& \winner{50.45} & 48.37
& \winner{4.23} & 4.22
& \winner{6.40} & 6.20
& \winner{5.83} & 6.15
& \winner{17.49} & 16.00
& \winfive{5/5}
& \winner{6.12} & 4.30
& \winner{1.21} & 0.89
& \winner{1.24} & 0.88
& \winner{5.38} & 6.20
& \winner{2.82} & 1.77
& \winfive{5/5} \\

W6
& 7.69 & \winner{8.04}
& \winner{1.07} & 1.04
& \winner{1.03} & 0.99
& \winner{8.94} & 10.17
& \winner{2.22} & 2.06
& \winfour{4/5}
& \winner{4.40} & 3.34
& \winner{0.95} & 0.74
& \winner{0.95} & 0.70
& 6.04 & \winner{5.42}
& \winner{1.86} & 1.61
& \winfour{4/5} \\

\specialrule{0.5pt}{2pt}{2pt}

% ====================== Panel B: Energy / Industrials ======================
\textbf{Window}
& \multicolumn{11}{c}{\textbf{Energy}}
& \multicolumn{11}{c}{\textbf{Industrials}} \\
\cmidrule(lr){2-12}
\cmidrule(lr){13-23}
& \multicolumn{2}{c}{\textbf{Ret. (\%) $\uparrow$}}
& \multicolumn{2}{c}{\textbf{Sharpe $\uparrow$}}
& \multicolumn{2}{c}{\textbf{Sortino $\uparrow$}}
& \multicolumn{2}{c}{\textbf{MDD (\%) $\downarrow$}}
& \multicolumn{2}{c}{\textbf{Calmar $\uparrow$}}
& \winhead{Win}
& \multicolumn{2}{c}{\textbf{Ret. (\%) $\uparrow$}}
& \multicolumn{2}{c}{\textbf{Sharpe $\uparrow$}}
& \multicolumn{2}{c}{\textbf{Sortino $\uparrow$}}
& \multicolumn{2}{c}{\textbf{MDD (\%) $\downarrow$}}
& \multicolumn{2}{c}{\textbf{Calmar $\uparrow$}}
& \winhead{Win} \\
\cmidrule(lr){2-3}
\cmidrule(lr){4-5}
\cmidrule(lr){6-7}
\cmidrule(lr){8-9}
\cmidrule(lr){10-11}
\cmidrule(lr){13-14}
\cmidrule(lr){15-16}
\cmidrule(lr){17-18}
\cmidrule(lr){19-20}
\cmidrule(lr){21-22}
& \multicolumn{1}{c}{\GIFThead{GIFT}} & \multicolumn{1}{c}{\basehead{Pure PPO}}
& \multicolumn{1}{c}{\GIFThead{GIFT}} & \multicolumn{1}{c}{\basehead{Pure PPO}}
& \multicolumn{1}{c}{\GIFThead{GIFT}} & \multicolumn{1}{c}{\basehead{Pure PPO}}
& \multicolumn{1}{c}{\GIFThead{GIFT}} & \multicolumn{1}{c}{\basehead{Pure PPO}}
& \multicolumn{1}{c}{\GIFThead{GIFT}} & \multicolumn{1}{c}{\basehead{Pure PPO}}
& \winhead{}
& \multicolumn{1}{c}{\GIFThead{GIFT}} & \multicolumn{1}{c}{\basehead{Pure PPO}}
& \multicolumn{1}{c}{\GIFThead{GIFT}} & \multicolumn{1}{c}{\basehead{Pure PPO}}
& \multicolumn{1}{c}{\GIFThead{GIFT}} & \multicolumn{1}{c}{\basehead{Pure PPO}}
& \multicolumn{1}{c}{\GIFThead{GIFT}} & \multicolumn{1}{c}{\basehead{Pure PPO}}
& \multicolumn{1}{c}{\GIFThead{GIFT}} & \multicolumn{1}{c}{\basehead{Pure PPO}}
& \winhead{} \\
\midrule

W1
& 23.88 & \winner{40.25}
& \winner{3.02} & 2.97
& \winner{3.54} & 3.34
& \winner{6.24} & 10.20
& \winner{8.58} & 8.46
& \winfour{4/5}
& \winner{15.89} & 12.38
& \winner{3.40} & 2.31
& \winner{3.79} & 2.69
& 4.11 & \winner{3.82}
& \winner{8.84} & 7.61
& \winfour{4/5} \\

W2
& \winner{20.38} & 17.44
& \winner{1.78} & 1.63
& \winner{1.93} & 1.71
& 9.94 & \winner{9.40}
& \winner{4.90} & 4.51
& \winfour{4/5}
& \winner{3.45} & 1.39
& \winner{0.65} & 0.34
& \winner{0.62} & 0.30
& 8.07 & \winner{5.77}
& \winner{1.14} & 0.71
& \winfour{4/5} \\

W3
& \winner{7.14} & 5.11
& \winner{0.63} & 0.52
& \winner{0.57} & 0.46
& 22.88 & \winner{20.96}
& \winner{1.05} & 0.87
& \winfour{4/5}
& \winner{-1.15} & -3.77
& \winner{-0.10} & -0.40
& \winner{-0.09} & -0.37
& \winner{13.48} & 15.46
& \winner{-0.11} & -0.49
& \winfive{5/5} \\

W4
& \winner{31.80} & 30.56
& \winner{2.14} & 2.12
& \winner{2.25} & 2.22
& 14.00 & \winner{13.69}
& \winner{5.20} & 5.13
& \winfour{4/5}
& \winner{12.86} & 6.75
& \winner{1.61} & 1.02
& \winner{1.68} & 1.01
& 13.78 & \winner{12.72}
& \winner{2.26} & 1.36
& \winfour{4/5} \\

W5
& \winner{-7.57} & -7.62
& \winner{-0.63} & -0.73
& \winner{-0.63} & -0.72
& 14.97 & \winner{13.61}
& \winner{-1.06} & -1.22
& \winfour{4/5}
& \winner{3.23} & 2.05
& \winner{0.76} & 0.46
& \winner{0.78} & 0.44
& \winner{4.28} & 4.78
& \winner{1.94} & 1.19
& \winfive{5/5} \\

W6
& \winner{-3.43} & -5.40
& \winner{-0.41} & -0.77
& \winner{-0.42} & -0.77
& \winner{11.44} & 12.12
& \winner{-0.62} & -1.02
& \winfive{5/5}
& \winner{9.16} & 5.81
& \winner{2.33} & 1.73
& \winner{2.47} & 1.76
& 7.53 & \winner{7.49}
& \winner{2.91} & 1.89
& \winfour{4/5} \\

\specialrule{0.5pt}{2pt}{2pt}

% ====================== Panel C: Light Mix / Heavy Mix ======================
\textbf{Window}
& \multicolumn{11}{c}{\textbf{Light Mix}}
& \multicolumn{11}{c}{\textbf{Heavy Mix}} \\
\cmidrule(lr){2-12}
\cmidrule(lr){13-23}
& \multicolumn{2}{c}{\textbf{Ret. (\%) $\uparrow$}}
& \multicolumn{2}{c}{\textbf{Sharpe $\uparrow$}}
& \multicolumn{2}{c}{\textbf{Sortino $\uparrow$}}
& \multicolumn{2}{c}{\textbf{MDD (\%) $\downarrow$}}
& \multicolumn{2}{c}{\textbf{Calmar $\uparrow$}}
& \winhead{Win}
& \multicolumn{2}{c}{\textbf{Ret. (\%) $\uparrow$}}
& \multicolumn{2}{c}{\textbf{Sharpe $\uparrow$}}
& \multicolumn{2}{c}{\textbf{Sortino $\uparrow$}}
& \multicolumn{2}{c}{\textbf{MDD (\%) $\downarrow$}}
& \multicolumn{2}{c}{\textbf{Calmar $\uparrow$}}
& \winhead{Win} \\
\cmidrule(lr){2-3}
\cmidrule(lr){4-5}
\cmidrule(lr){6-7}
\cmidrule(lr){8-9}
\cmidrule(lr){10-11}
\cmidrule(lr){13-14}
\cmidrule(lr){15-16}
\cmidrule(lr){17-18}
\cmidrule(lr){19-20}
\cmidrule(lr){21-22}
& \multicolumn{1}{c}{\GIFThead{GIFT}} & \multicolumn{1}{c}{\basehead{Pure PPO}}
& \multicolumn{1}{c}{\GIFThead{GIFT}} & \multicolumn{1}{c}{\basehead{Pure PPO}}
& \multicolumn{1}{c}{\GIFThead{GIFT}} & \multicolumn{1}{c}{\basehead{Pure PPO}}
& \multicolumn{1}{c}{\GIFThead{GIFT}} & \multicolumn{1}{c}{\basehead{Pure PPO}}
& \multicolumn{1}{c}{\GIFThead{GIFT}} & \multicolumn{1}{c}{\basehead{Pure PPO}}
& \winhead{}
& \multicolumn{1}{c}{\GIFThead{GIFT}} & \multicolumn{1}{c}{\basehead{Pure PPO}}
& \multicolumn{1}{c}{\GIFThead{GIFT}} & \multicolumn{1}{c}{\basehead{Pure PPO}}
& \multicolumn{1}{c}{\GIFThead{GIFT}} & \multicolumn{1}{c}{\basehead{Pure PPO}}
& \multicolumn{1}{c}{\GIFThead{GIFT}} & \multicolumn{1}{c}{\basehead{Pure PPO}}
& \multicolumn{1}{c}{\GIFThead{GIFT}} & \multicolumn{1}{c}{\basehead{Pure PPO}}
& \winhead{} \\
\midrule

W1
& \winner{2.41} & -1.78
& \winner{0.45} & -0.14
& \winner{0.45} & -0.14
& \winner{10.04} & 11.93
& \winner{0.70} & -0.22
& \winfive{5/5}
& \winner{14.26} & 7.72
& \winner{1.73} & 0.84
& \winner{1.83} & 0.88
& \winner{9.08} & 12.75
& \winner{3.77} & 1.66
& \winfive{5/5} \\

W2
& \winner{12.16} & 12.13
& \winner{2.11} & 2.08
& 2.10 & \winner{2.12}
& \winner{6.74} & 7.65
& \winner{4.26} & 3.75
& \winfour{4/5}
& 19.23 & \winner{23.16}
& 2.41 & \winner{2.71}
& 2.63 & \winner{3.30}
& 6.95 & \winner{5.81}
& 6.38 & \winner{9.02}
& \winzero{0/5} \\

W3
& \winner{-13.61} & -32.70
& \winner{-1.25} & -2.05
& \winner{-1.13} & -1.72
& \winner{21.23} & 35.98
& \winner{-1.51} & -2.41
& \winfive{5/5}
& \winner{-6.90} & -15.83
& \winner{-0.64} & -0.90
& \winner{-0.61} & -0.85
& \winner{14.96} & 26.44
& \winner{-0.98} & -1.30
& \winfive{5/5} \\

W4
& \winner{-9.17} & -23.46
& \winner{-0.72} & -2.30
& \winner{-0.77} & -2.42
& \winner{13.70} & 26.35
& \winner{-1.43} & -2.32
& \winfive{5/5}
& \winner{-4.17} & -10.04
& \winner{-0.28} & -0.97
& \winner{-0.30} & -0.92
& \winner{13.40} & 17.03
& \winner{-0.53} & -1.34
& \winfive{5/5} \\

W5
& \winner{20.40} & 13.62
& \winner{2.71} & 1.99
& \winner{2.98} & 2.32
& \winner{10.05} & 11.33
& \winner{4.63} & 2.85
& \winfive{5/5}
& \winner{17.01} & 11.58
& \winner{2.57} & 1.60
& \winner{2.96} & 1.62
& \winner{4.79} & 8.38
& \winner{8.20} & 3.36
& \winfive{5/5} \\

W6
& \winner{5.80} & 5.28
& \winner{1.11} & 0.79
& \winner{1.07} & 0.77
& \winner{8.89} & 12.28
& \winner{1.65} & 1.16
& \winfive{5/5}
& \winner{4.14} & 2.86
& \winner{0.72} & 0.55
& \winner{0.76} & 0.58
& 12.02 & \winner{11.87}
& \winner{0.92} & 0.67
& \winfour{4/5} \\

\bottomrule
\end{tabular}
\end{adjustbox}
\end{table*}
\subsection{Out-of-Sample Portfolio Performance (RQ1)}
\label{sec:main_performance_comparison}

\paragraph{\bfseries\itshape Overall numerical performance.}
We first evaluate whether the state-reward interface selected by GIFT improves PPO under the controlled evaluation protocol.
Across six portfolio panels and six rolling windows, GIFT-enhanced PPO is compared with Pure PPO using the same policy architecture, hyperparameters, and chronological split.
The only difference is the state-reward interface: Pure PPO keeps the original state and reward, while GIFT uses the interface fixed after the offline design stage.
Table~\ref{tab:main_performance_all_sectors} reports Return, Sharpe, Sortino, MDD, and Calmar.
We count a panel--window cell as a win if GIFT outperforms Pure PPO on at least four of the five metrics.

As shown in Table~\ref{tab:main_performance_all_sectors}, GIFT wins 160 of 180 metric-level comparisons, and 35 of 36 panel--window cells reach at least 4/5 wins.
The gains are consistent across Technology, Healthcare, Energy, Industrials, and Light Mix, while Heavy Mix improves in five of six windows.
The only non-winning cell is Heavy Mix in W2, where Pure PPO performs better on all five metrics; even in this case, GIFT remains competitive with a Sharpe ratio of 2.41 and an MDD of 6.95\%.

The improvements are more pronounced in stressed windows such as W3 and W4.
In these periods, GIFT often reduces drawdown and improves Sharpe, Sortino, and Calmar while maintaining or improving return.
This pattern suggests that the selected state-reward interface helps PPO learn a more stable return-risk trade-off, rather than merely increasing portfolio exposure. \textbf{These results indicate that GIFT improves out-of-sample performance mainly through stronger risk-adjusted behavior under the same PPO evaluation protocol.}

\begin{figure}[t]
    \centering
    \includegraphics[width=0.88\columnwidth]{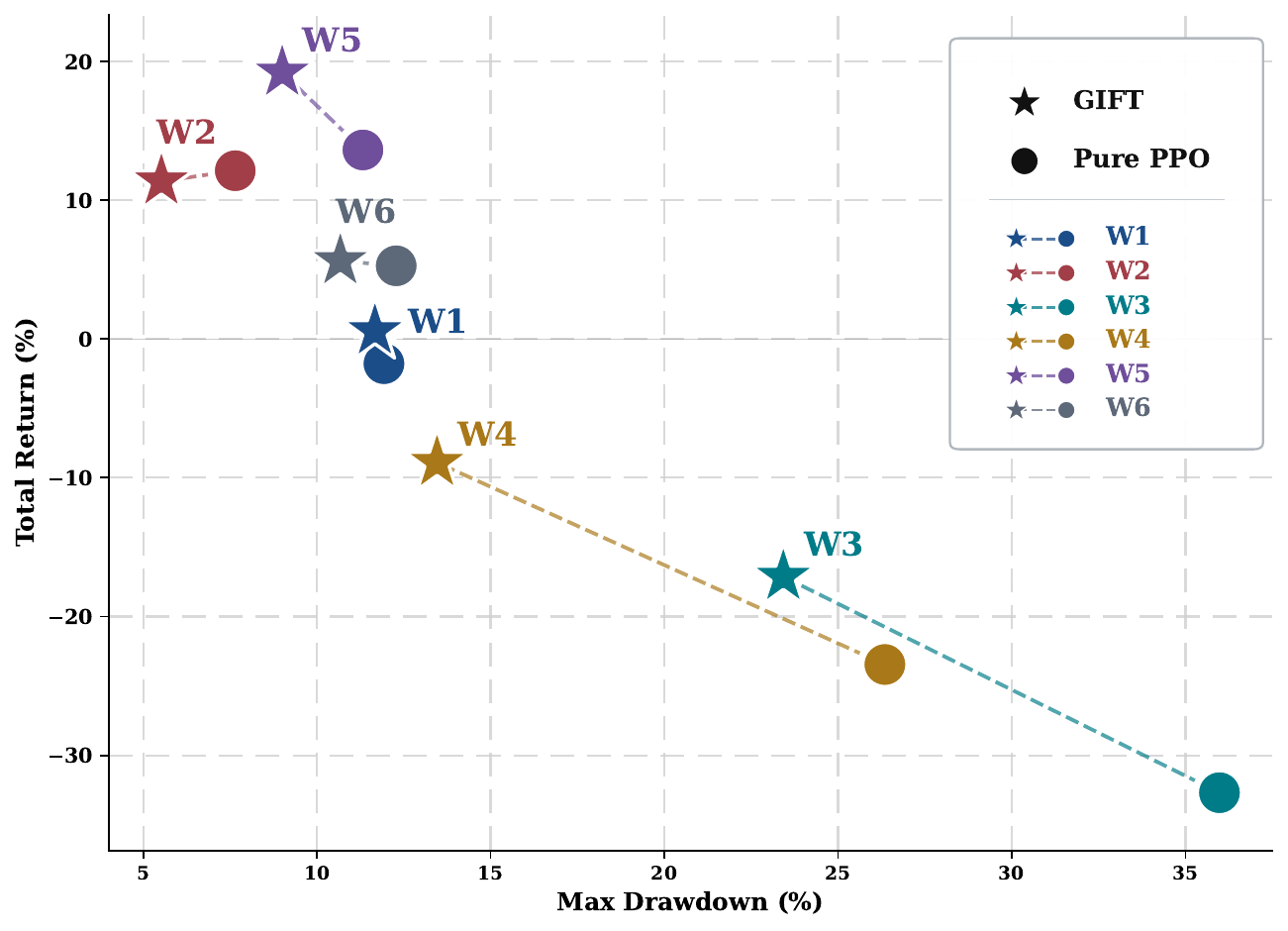}
    \caption{
    Risk--return profile on the Light Mix panel across rolling windows.
    Points closer to the upper-left indicate higher return with lower maximum drawdown.
    }
    \label{fig:risk_return_profile}
\end{figure}

\paragraph{\bfseries\itshape Portfolio behavior and risk allocation.}
We further inspect the learned portfolio behavior on the Light Mix panel.
Figure~\ref{fig:risk_return_profile} plots total return against maximum drawdown across rolling windows.
GIFT is closer to the upper-left region in most windows, indicating higher return with lower drawdown.
The separation is clearest in W3 and W4, where Pure PPO suffers larger drawdowns and lower returns.

\begin{figure}[t]
    \centering
    \includegraphics[width=0.88\columnwidth]{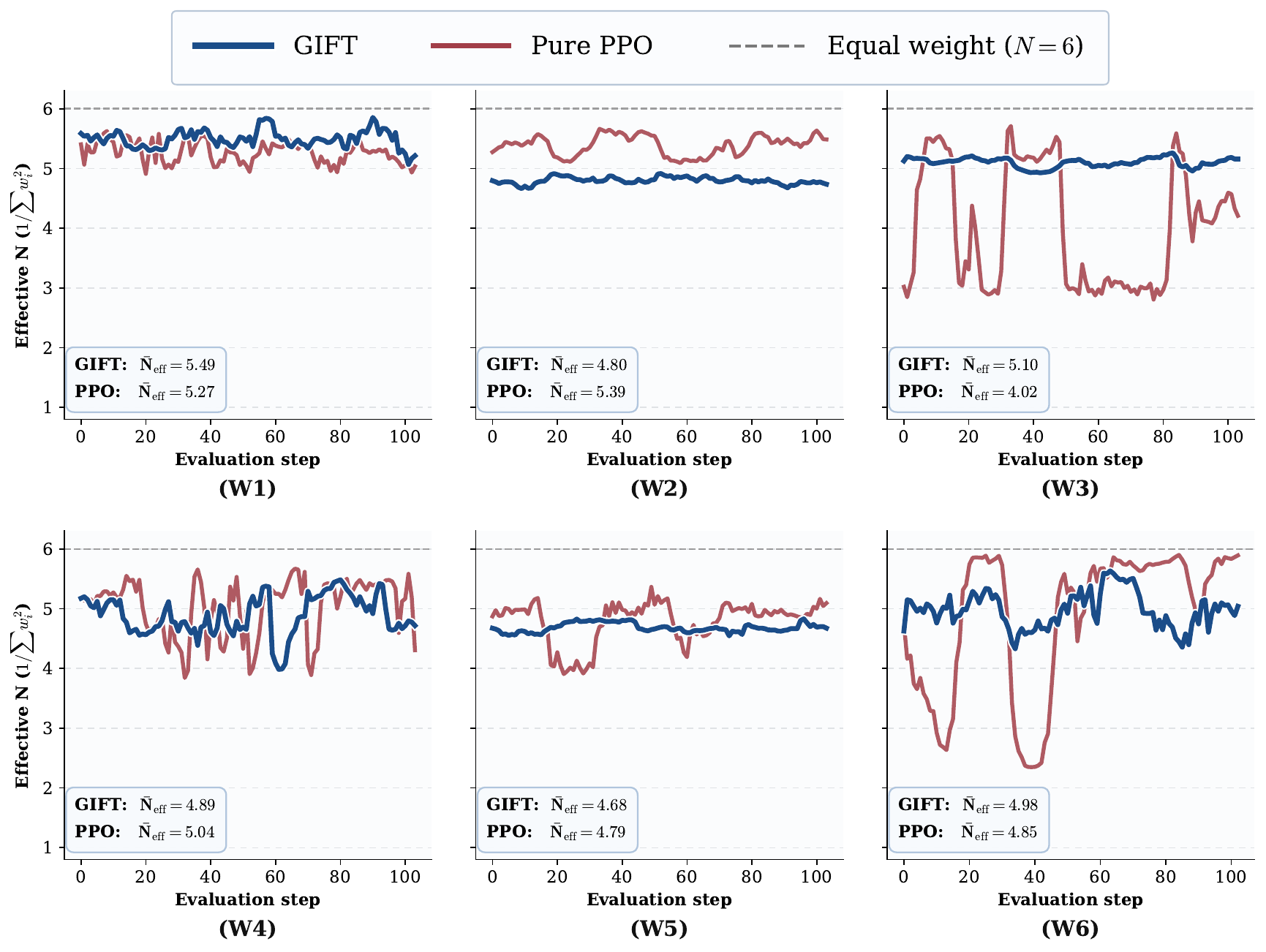}
    \caption{
    Effective diversification on the Light Mix panel, measured by
    \(N_{\mathrm{eff}}=1/\sum_i w_i^2\).
    Higher values indicate less concentrated allocations.
    }
    \label{fig:effective_diversification}
    \vspace{-4mm}
\end{figure}

Figure~\ref{fig:effective_diversification} provides a complementary view of allocation behavior.
We measure diversification by \(N_{\mathrm{eff}}=1/\sum_i w_i^2\), where larger values indicate less concentrated allocations.
GIFT keeps \(N_{\mathrm{eff}}\) more stable across evaluation steps, whereas Pure PPO shows sharper drops in several windows, especially W3 and W6.
These drops indicate temporary concentration in fewer assets and are consistent with the larger drawdowns observed for Pure PPO. \textbf{On Light Mix, the performance gains of GIFT are associated with more stable diversification and better drawdown control.}

\subsection{State-reward Interface Variants (RQ2)}
\label{sec:design_mechanism}

We evaluate how different state-reward interface designs affect PPO training feedback and cross-window performance.

\paragraph{\bfseries\itshape Controlled interface variants.}
Tables~\ref{tab:design_variants} and~\ref{tab:variant_stability} summarize the compared variants under the same PPO backbone and rolling-window protocol.
Fixed Indicators, Fixed Reward, and Fixed State-Reward improve some windows but are less stable under regime shifts, while Free-form LLM remains brittle.
GIFT achieves 29/30 metric-level wins and at least 4/5 wins in every window, suggesting that guided LLM-generated interfaces are more reliable than fixed designs or unconstrained generation.

\begin{table}[t]
\centering
\scriptsize
\setlength{\tabcolsep}{3.5pt}
\renewcommand{\arraystretch}{1.12}
\caption{
state-reward interface variants under the same PPO backbone.
}
\label{tab:design_variants}
\resizebox{\columnwidth}{!}{
\begin{tabular}{lll}
\toprule
\textbf{Method}
& \textbf{State Design}
& \textbf{Reward Design} \\
\midrule
Pure PPO
& Raw OHLCV
& Short-horizon return \\

Fixed Indicators
& Handcrafted financial indicators
& Short-horizon return \\

Fixed Reward
& Raw OHLCV
& Fixed risk-rule reward \\

Fixed State-Reward
& Handcrafted financial indicators
& Fixed risk-rule reward \\

Free-form LLM
& Unguided LLM-generated features
& Unguided LLM-generated reward \\

GIFT
& Factor-guided LLM state enhancement
& Risk-rule-guided LLM reward shaping \\
\bottomrule
\end{tabular}
}
\end{table}

\begin{table}[t]
\centering
\scriptsize
\setlength{\tabcolsep}{3.5pt}
\renewcommand{\arraystretch}{1.12}
\caption{
Cross-window stability of state-reward interface variants.
W1--W6 report metric-level wins over five metrics; full results are in Appendix~\ref{app:interface_variant_details}.
}
\label{tab:variant_stability}
\resizebox{\columnwidth}{!}{
\begin{tabular}{lcccccccc}
\toprule
\textbf{Method}
& \textbf{Total}
& \textbf{Win Rate}
& \textbf{W1}
& \textbf{W2}
& \textbf{W3}
& \textbf{W4}
& \textbf{W5}
& \textbf{W6} \\
\midrule

Fixed Indicators
& 18/30 & 60.0\%
& \winfour{4}
& \wintwo{2}
& \winfive{5}
& \wintwo{2}
& \winzero{0}
& \winfive{5} \\

Fixed Reward
& 20/30 & 66.7\%
& \winfive{5}
& \wintwo{2}
& \winfive{5}
& \winthree{3}
& \winzero{0}
& \winfive{5} \\

Fixed State-Reward
& 22/30 & 73.3\%
& \winfive{5}
& \winthree{3}
& \winfive{5}
& \winfour{4}
& \winzero{0}
& \winfive{5} \\

Free-form LLM
& 19/30 & 63.3\%
& \winfive{5}
& \winzero{0}
& \winfive{5}
& \winfour{4}
& \winfive{5}
& \winzero{0} \\

\textbf{GIFT}
& \textbf{29/30} & \textbf{96.7\%}
& \winfive{5}
& \winfour{4}
& \winfive{5}
& \winfive{5}
& \winfive{5}
& \winfive{5} \\

\bottomrule
\end{tabular}
}
\end{table}

\begin{figure}[t]
    \centering
    \includegraphics[width=\columnwidth]{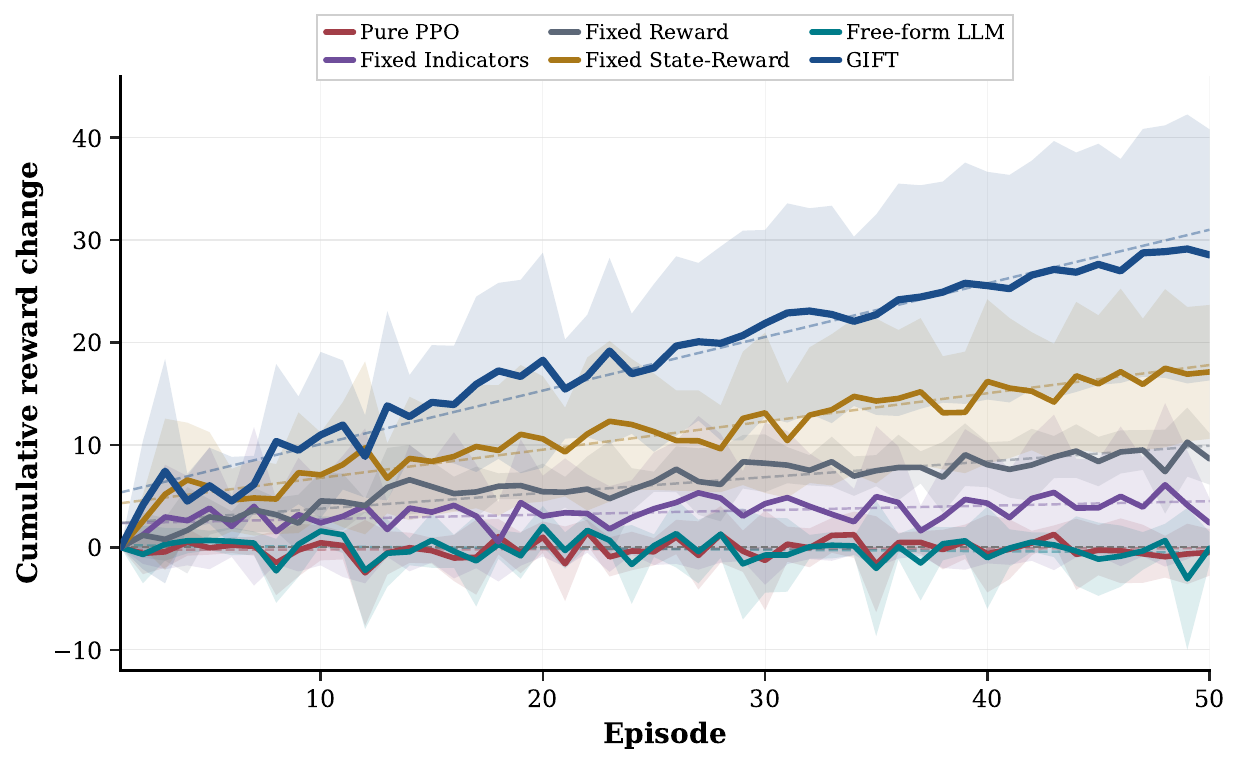}
\caption{
Cumulative reward change during PPO training.
Shaded regions denote variability.
}
    \label{fig:cumulative_reward_change}
\vspace{-4mm}
\end{figure}

\paragraph{\bfseries\itshape Training-signal diagnostics.}
Figure~\ref{fig:cumulative_reward_change} and Table~\ref{tab:variant_signal_quality} further examine PPO training feedback.
GIFT shows a more sustained upward reward trajectory and obtains the strongest SNR, autocorrelation, trend, and reward--return alignment.
Free-form LLM instead yields weak signals, including negative autocorrelation and trend.
\textbf{Overall, GIFT's advantage comes from guided offline interface design, not from fixed feature engineering, fixed reward shaping, or unconstrained LLM generation.}

\begin{table}[t]
\centering
\scriptsize
\setlength{\tabcolsep}{3.8pt}
\renewcommand{\arraystretch}{1.12}
\caption{
Reward-signal quality across state-reward interface variants.
SNR, Trend, and Alignment measure signal clarity, reward-curve slope, and correlation with future portfolio return.
}
\label{tab:variant_signal_quality}
\resizebox{\columnwidth}{!}{
\begin{tabular}{lcccc}
\toprule
\textbf{Method}
& \textbf{SNR$\uparrow$}
& \textbf{Autocorr.}
& \textbf{Trend$\uparrow$}
& \textbf{Alignment$\uparrow$} \\
\midrule
Pure PPO
& 0.013 & 0.051 & 0.003 & 0.174 \\

Fixed Indicators
& 0.020 & 0.153 & 0.009 & 0.335 \\

Fixed Reward
& 0.015 & 0.123 & 0.005 & 0.206 \\

Fixed State-Reward
& 0.009 & 0.086 & 0.003 & 0.125 \\

Free-form LLM
& 0.006 & $-$0.037 & $-$0.012 & 0.042 \\

\textbf{GIFT}
& \textbf{0.181} & \textbf{0.733} & \textbf{0.338} & \textbf{0.849} \\
\bottomrule
\end{tabular}
}
\end{table}

\subsection{Component Ablation (RQ3)}
\label{sec:component_ablation}

\paragraph{\bfseries\itshape Main component removal.}
Table~\ref{tab:ablation_main} reports average changes over six Technology-sector rolling windows under the same PPO backbone and evaluation protocol.
Removing DGR gives the largest degradation, reducing Sharpe by 0.393, Sortino by 0.400, Calmar by 0.513, and Return by 3.10 percentage points.
Removing FSE and RRS also weakens performance, suggesting that both state enhancement and reward shaping contribute to the selected interface.

\begin{table}[t]
\centering
\scriptsize
\setlength{\tabcolsep}{4.2pt}
\renewcommand{\arraystretch}{1.12}
\caption{
Average ablation results over six rolling windows.
$\Delta$ is relative to Full GIFT; MDD and Return changes are reported in percentage points.
See Appendix~\ref{app:ablation_details}.
}
\label{tab:ablation_main}
\resizebox{\columnwidth}{!}{
\begin{tabular}{lccccc}
\toprule
\textbf{Method}
& \textbf{$\Delta$Sharpe$\uparrow$}
& \textbf{$\Delta$Sortino$\uparrow$}
& \textbf{$\Delta$MDD$\downarrow$}
& \textbf{$\Delta$Calmar$\uparrow$}
& \textbf{$\Delta$Return$\uparrow$} \\
\midrule

w/o FSE
& $-$0.129
& $-$0.122
& +1.30 pp
& $-$0.209
& $-$1.56 pp \\

w/o RRS
& $-$0.129
& $-$0.213
& +3.05 pp
& +0.062
& $-$0.57 pp \\

w/o DGR
& $-$0.393
& $-$0.400
& +2.56 pp
& $-$0.513
& $-$3.10 pp \\

w/o IC/SHAP
& $-$0.257
& $-$0.204
& +3.59 pp
& $-$0.345
& $-$1.92 pp \\

\bottomrule
\end{tabular}
}
\end{table}

\paragraph{\bfseries\itshape Diagnostic feedback ablation.}
Removing IC/SHAP diagnostics produces the largest MDD increase of 3.59 percentage points and also reduces Sharpe, Sortino, Calmar, and Return.
\textbf{Overall, GIFT benefits from state-reward interface design and diagnostic feedback, with DGR contributing most in this setting.}

\subsection{Adaptive Design Selection and Generated Semantics (RQ4)}
\label{sec:semantic_analysis}

We examine whether GIFT selects different, interpretable state-reward interfaces across rolling market windows.
We track the selected factors, risk-rule subsets, intrinsic-reward patterns, and refinement rounds.

\vspace{-2mm}

\paragraph{\bfseries\itshape Window-level interface selection.}
Table~\ref{tab:window_adaptive_design} summarizes the selected interface in each window.
The selected designs vary across regimes rather than collapsing to a single template.
Bull or recovery windows tend to select diversification or multi-factor designs, while stressed or post-shock windows place more emphasis on mean reversion, state-reward coupling, or defensive risk control.
Since the selected interface is frozen before evaluation, these differences come from historical design-window diagnostics rather than evaluation-window feedback. Appendix~\ref{app:case_studies} provides operation-frequency statistics and qualitative examples.

\textbf{Overall, GIFT selects regime-conditioned interfaces during offline design, while keeping the selected interface fixed during evaluation.}

\begin{table}[!htbp]
\centering
\scriptsize
\setlength{\tabcolsep}{4pt}
\renewcommand{\arraystretch}{1.15}
\caption{
Window-level selected interfaces.
Each interface is selected by rollout score before evaluation and then kept fixed.
}
\label{tab:window_adaptive_design} 
\resizebox{\columnwidth}{!}{%
\begin{tabular}{c c l c r}
\toprule
\textbf{Win.} & \textbf{Regime} & \textbf{Dominant design pattern} & \textbf{Iter.} & \textbf{Train Sharpe} \\
\midrule
W1 & \rUp & \textsc{Diversification}\,+ turnover ctrl & \iterDot{1} & $+1.250$ \\
W2 & \rUp & \textsc{Multi-factor}\,+ momentum         & \iterDot{2} & $+1.823$ \\
W3 & \rDn & \textsc{Mean reversion}\,+ S--R coupling  & \iterDot{3} & $+1.388$ \\
W4 & \rMx & \textsc{Risk-aware stab.}                 & \iterDot{2} & $+0.788$ \\
W5 & \rMx & \textsc{Multi-factor}\,+ tail-risk        & \iterDot{1} & $-0.548$ \\
W6 & \rV  & \textsc{Defensive risk ctrl}              & \iterDot{2} & $-0.593$ \\
\bottomrule
\end{tabular}}
\end{table}

\subsection{Transferability and Robustness Analysis (RQ5)}
\label{sec:algorithmic_transfer}

\paragraph{\bfseries\itshape Backbone transfer.}
We replace PPO with A2C while keeping the GIFT interface-search process unchanged.
As shown in Table~\ref{tab:GIFT_a2c_transfer}, GIFT+A2C wins 28 of 30 metric-level comparisons against pure A2C, suggesting that the generated interface is not tightly tied to PPO-specific optimization.
\begin{table}[t]
\centering
\tiny
\setlength{\tabcolsep}{2.4pt}
\renewcommand{\arraystretch}{1.08}
\caption{Transfer robustness under A2C.}
\label{tab:GIFT_a2c_transfer}
\resizebox{\columnwidth}{!}{
\begin{tabular}{lccccccccccc}
\toprule
\multicolumn{12}{c}{\textbf{GIFT+A2C vs. Pure A2C}} \\
\midrule
\textbf{Window}
& \multicolumn{2}{c}{\textbf{Return}}
& \multicolumn{2}{c}{\textbf{Sharpe}}
& \multicolumn{2}{c}{\textbf{Sortino}}
& \multicolumn{2}{c}{\textbf{MDD}}
& \multicolumn{2}{c}{\textbf{Calmar}}
& \winhead{Win} \\
\cmidrule(lr){2-3}
\cmidrule(lr){4-5}
\cmidrule(lr){6-7}
\cmidrule(lr){8-9}
\cmidrule(lr){10-11}
&
\GIFThead{GIFT+A2C} & \basehead{Pure A2C}
& \GIFThead{GIFT+A2C} & \basehead{Pure A2C}
& \GIFThead{GIFT+A2C} & \basehead{Pure A2C}
& \GIFThead{GIFT+A2C} & \basehead{Pure A2C}
& \GIFThead{GIFT+A2C} & \basehead{Pure A2C}
& \winhead{} \\
\midrule

W1
& +0.33 & $-$5.71
& +0.134 & $-$0.365
& +0.130 & $-$0.380
& 11.66 & 18.55
& +0.208 & $-$0.554
& \wincell{5/5} \\

W2
& +15.35 & +15.01
& +2.423 & +2.297
& +2.492 & +2.232
& 7.26 & 8.18
& +4.920 & +4.286
& \wincell{5/5} \\

W3
& $-$19.90 & $-$38.81
& $-$1.375 & $-$2.011
& $-$1.227 & $-$1.742
& 28.12 & 42.52
& $-$1.695 & $-$2.461
& \wincell{5/5} \\

W4
& $-$19.71 & $-$21.59
& $-$1.754 & $-$2.301
& $-$1.845 & $-$2.415
& 24.11 & 24.70
& $-$2.041 & $-$2.265
& \wincell{5/5} \\

W5
& +19.90 & +18.60
& +2.817 & +2.251
& +3.117 & +2.237
& 9.35 & 11.25
& +4.844 & +3.841
& \wincell{5/5} \\

W6
& +5.60 & +7.31
& +1.236 & +1.204
& +1.208 & +1.180
& 7.60 & 8.98
& +1.837 & +2.052
& \wincell{3/5} \\

\midrule
\textbf{Summary}
& \multicolumn{10}{c}{5/6 windows reach a 5/5 metric sweep; the remaining window improves 3/5 metrics.}
& \wincell{28/30} \\
\bottomrule
\end{tabular}
}
\end{table}

\begin{figure}[t]
    \centering
    \includegraphics[width=\columnwidth]{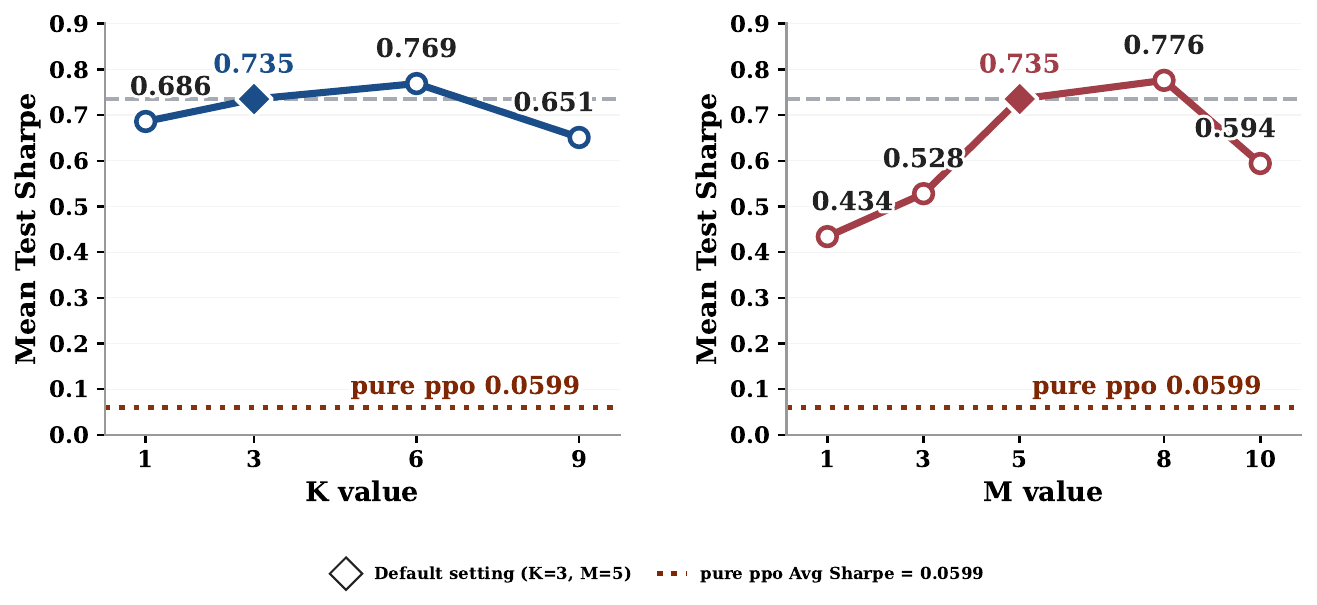}
    \caption{
    Sensitivity to offline design settings.
    Diamonds mark the default \((K=3,M=5)\); dashed lines denote Pure PPO.
    }
    \label{fig:hyperparameter_sensitivity}
\end{figure}

\paragraph{\bfseries\itshape Offline design sensitivity.}
We further vary the number of candidate interfaces \(K\) and refinement rounds \(M\).
As shown in Figure~\ref{fig:hyperparameter_sensitivity}, GIFT remains above the Pure PPO reference across nearby offline design settings, indicating that the improvement is not driven by a single search-budget choice.
Additional seed and stronger-baseline checks are reported in Appendices~\ref{app:multiseed_robustness} and~\ref{app:additional_train_plus_adapt_ppo_baseline_robustness}.
\textbf{These checks suggest that the gains are not tied to a single PPO implementation, search budget, random seed, or baseline configuration.}

\vspace{-5pt}
% \section{Conclusion}

\section{Conclusion}

We proposed GIFT, an LLM-guided framework for state-reward interface design in financial reinforcement learning. 
Instead of using the LLM as a direct trading agent, GIFT uses it as a constrained financial-knowledge-guided interface designer, combining factor-guided state enhancement, risk-rule-guided reward shaping, and diagnostic-guided refinement. 
Across rolling-window experiments, GIFT improves PPO's learning signals and out-of-sample risk-adjusted performance under the same evaluation protocol. 
Further variant, ablation, and robustness analyses suggest that the gains mainly come from guided offline interface design, rather than fixed designs or unconstrained LLM generation. 
These results support a more controlled role for LLMs in portfolio RL: designing the learning interface while leaving trading decisions to the RL policy.
\vspace{-5pt}

\clearpage

\section*{Limitations}

This work relies on historical daily OHLCV data and rolling-window backtesting, which cannot fully reflect real trading frictions such as liquidity, execution delay, market impact, taxes, or time-varying costs.
Its generalization to other assets, frequencies, constraints, DRL backbones, and design libraries remains for future study.
\vspace{-5pt}
\section*{Ethical Considerations}

This paper is for research only and does not provide investment advice or financial guarantees.
GIFT uses LLMs only as constrained state-reward design generators; any real deployment would require additional risk control, regulatory compliance, human oversight, and realistic execution validation.

\bibliography{main}

@ARTICLE{moody2001direct,
  author={Moody, J. and Saffell, M.},
  journal={IEEE Transactions on Neural Networks}, 
  title={Learning to trade via direct reinforcement}, 
  year={2001},
  volume={12},
  number={4},
  pages={875-889},
  doi={10.1109/72.935097}}

@misc{jiang2017portfolio,
      title={A Deep Reinforcement Learning Framework for the Financial Portfolio Management Problem}, 
      author={Zhengyao Jiang and Dixing Xu and Jinjun Liang},
      year={2017},
      eprint={1706.10059},
      archivePrefix={arXiv},
      primaryClass={q-fin.CP},
      url={https://arxiv.org/abs/1706.10059}, 
}

@article{sun2023rlqt,
author = {Sun, Shuo and Wang, Rundong and An, Bo},
title = {Reinforcement Learning for Quantitative Trading},
year = {2023},
issue_date = {June 2023},
publisher = {Association for Computing Machinery},
address = {New York, NY, USA},
volume = {14},
number = {3},
issn = {2157-6904},
url = {https://doi.org/10.1145/3582560},
doi = {10.1145/3582560},
journal = {ACM Trans. Intell. Syst. Technol.},
month = mar,
articleno = {44},
numpages = {29}
}

@article{hambly2023rlfinance,
author = {Hambly, Ben and Xu, Renyuan and Yang, Huining},
title = {Recent advances in reinforcement learning in finance},
journal = {Mathematical Finance},
volume = {33},
number = {3},
pages = {437-503},
doi = {https://doi.org/10.1111/mafi.12382},
url = {https://onlinelibrary.wiley.com/doi/abs/10.1111/mafi.12382},
eprint = {https://onlinelibrary.wiley.com/doi/pdf/10.1111/mafi.12382},
year = {2023}
}

@article{pippas2025rlqfsurvey,
author = {Pippas, Nikolaos and Ludvig, Elliot A. and Turkay, Cagatay},
title = {The Evolution of Reinforcement Learning in Quantitative Finance: A Survey},
year = {2025},
issue_date = {November 2025},
publisher = {Association for Computing Machinery},
address = {New York, NY, USA},
volume = {57},
number = {11},
issn = {0360-0300},
url = {https://doi.org/10.1145/3733714},
doi = {10.1145/3733714},
journal = {ACM Comput. Surv.},
month = jun,
articleno = {295},
numpages = {51}
}

@article{rezaei2025drlpm,
  title = {A Taxonomy of Literature Reviews and Experimental Study of Deepreinforcement Learning in Portfolio Management},
  author = {Rezaei, Mohadese and {Nezamabadi-Pour}, Hossein},
  year = 2025,
  month = jan,
  journal = {Artificial Intelligence Review},
  volume = {58},
  number = {3},
  pages = {94},
  issn = {1573-7462},
  doi = {10.1007/s10462-024-11066-w}
}

@inproceedings{liu2022finrlmeta,
author = {Liu, Xiao-Yang and Xia, Ziyi and Rui, Jingyang and Gao, Jiechao and Yang, Hongyang and Zhu, Ming and Wang, Christina Dan and Wang, Zhaoran and Guo, Jian},
title = {FinRL-meta: market environments and benchmarks for data-driven financial reinforcement learning},
year = {2022},
isbn = {9781713871088},
publisher = {Curran Associates Inc.},
address = {Red Hook, NY, USA},
booktitle = {Proceedings of the 36th International Conference on Neural Information Processing Systems},
articleno = {134},
numpages = {15},
location = {New Orleans, LA, USA},
series = {NIPS '22}
}

@misc{schulman2017ppo,
      title={Proximal Policy Optimization Algorithms}, 
      author={John Schulman and Filip Wolski and Prafulla Dhariwal and Alec Radford and Oleg Klimov},
      year={2017},
      eprint={1707.06347},
      archivePrefix={arXiv},
      primaryClass={cs.LG},
      url={https://arxiv.org/abs/1707.06347}, 
}

@article{goluza2024positional,
  title = {Deep Reinforcement Learning with Positional Context for Intraday Trading},
  author = {Golu{\v z}a, Sven and Kova{\v c}evi{\'c}, Tomislav and Bauman, Tessa and Kostanj{\v c}ar, Zvonko},
  year = 2024,
  month = oct,
  journal = {Evolving Systems},
  volume = {15},
  number = {5},
  pages = {1865--1880},
  issn = {1868-6486},
  doi = {10.1007/s12530-024-09593-6}
}

@misc{liang2018adversarial,
      title={Adversarial Deep Reinforcement Learning in Portfolio Management}, 
      author={Zhipeng Liang and Hao Chen and Junhao Zhu and Kangkang Jiang and Yanran Li},
      year={2018},
      eprint={1808.09940},
      archivePrefix={arXiv},
      primaryClass={q-fin.PM},
      url={https://arxiv.org/abs/1808.09940}, 
}

@inproceedings{ng1999rewardshaping,
author = {Ng, Andrew Y. and Harada, Daishi and Russell, Stuart J.},
title = {Policy Invariance Under Reward Transformations: Theory and Application to Reward Shaping},
year = {1999},
isbn = {1558606122},
publisher = {Morgan Kaufmann Publishers Inc.},
address = {San Francisco, CA, USA},
booktitle = {Proceedings of the Sixteenth International Conference on Machine Learning},
pages = {278–287},
numpages = {10},
series = {ICML '99}
}

@INPROCEEDINGS{goluza2024imitationreward,
  author={Goluža, Sven and Kovačević, Tomislav and Begušić, Stjepan and Kostanjčar, Zvonko},
  booktitle={2024 IEEE International Conference on Big Data (BigData)}, 
  title={Robot See, Robot Do: Imitation Reward for Noisy Financial Environments}, 
  year={2024},
  volume={},
  number={},
  pages={4884-4891},
  doi={10.1109/BigData62323.2024.10825427}}

@inproceedings{karzanov2025regretreward,
author = {Karzanov, Daniil and Garz\'{o}n, Rub\'{e}n and Terekhov, Mikhail and Gulcehre, Caglar and Raffinot, Thomas and Detyniecki, Marcin},
title = {Regret-Optimized Portfolio Enhancement through Deep Reinforcement Learning and Future Looking Rewards},
year = {2025},
isbn = {9798400722202},
publisher = {Association for Computing Machinery},
address = {New York, NY, USA},
url = {https://doi.org/10.1145/3768292.3770340},
doi = {10.1145/3768292.3770340},
booktitle = {Proceedings of the 6th ACM International Conference on AI in Finance},
pages = {890–897},
numpages = {8},
location = {
},
series = {ICAIF '25}
}

@article{wu2023bloomberggpt,
  title={BloombergGPT: A Large Language Model for Finance},
  author={Shijie Wu and Ozan Irsoy and Steven Lu and Vadim Dabravolski and Mark Dredze and Sebastian Gehrmann and Prabhanjan Kambadur and David Stuart Rosenberg and Gideon Mann},
  journal={ArXiv},
  year={2023},
  volume={abs/2303.17564},
  url={https://api.semanticscholar.org/CorpusID:257833842}
}

@article{yang2023fingpt,
  title={FinGPT: Open-Source Financial Large Language Models},
  author={Hongyang Yang and Xiao-Yang Liu and Chris Wang},
  journal={ArXiv},
  year={2023},
  volume={abs/2306.06031},
  url={https://api.semanticscholar.org/CorpusID:259129734}
}

@misc{nie2024finllmsurvey,
      title={A Survey of Large Language Models for Financial Applications: Progress, Prospects and Challenges}, 
      author={Yuqi Nie and Yaxuan Kong and Xiaowen Dong and John M. Mulvey and H. Vincent Poor and Qingsong Wen and Stefan Zohren},
      year={2024},
      eprint={2406.11903},
      archivePrefix={arXiv},
      primaryClass={q-fin.GN},
      url={https://arxiv.org/abs/2406.11903}, 
}

@inproceedings{xie2024finben,
author = {Xie, Qianqian and Han, Weiguang and Chen, Zhengyu and Xiang, Ruoyu and Zhang, Xiao and He, Yueru and Xiao, Mengxi and Li, Dong and Dai, Yongfu and Feng, Duanyu and Xu, Yijing and Kang, Haoqiang and Kuang, Ziyan and Yuan, Chenhan and Yang, Kailai and Luo, Zheheng and Zhang, Tianlin and Liu, Zhiwei and Xiong, Guojun and Deng, Zhiyang and Jiang, Yuechen and Yao, Zhiyuan and Li, Haohang and Yu, Yangyang and Hu, Gang and Huang, Jiajia and Liu, Xiao-Yang and Lopez-Lira, Alejandro and Wang, Benyou and Lai, Yanzhao and Wang, Hao and Peng, Min and Ananiadou, Sophia and Huang, Jimin},
title = {FinBen: a holistic financial benchmark for large language models},
year = {2024},
isbn = {9798331314385},
publisher = {Curran Associates Inc.},
address = {Red Hook, NY, USA},
booktitle = {Proceedings of the 38th International Conference on Neural Information Processing Systems},
articleno = {3033},
numpages = {28},
location = {Vancouver, BC, Canada},
series = {NIPS '24}
}

@inproceedings{yu2024fincon,
author = {Yu, Yangyang and Yao, Zhiyuan and Li, Haohang and Deng, Zhiyang and Jiang, Yuechen and Cao, Yupeng and Chen, Zhi and Suchow, Jordan W. and Cui, Zhenyu and Liu, Rong and Xu, Zhaozhuo and Zhang, Denghui and Subbalakshmi, Koduvayur and Xiong, Guojun and He, Yueru and Huang, Jimin and Li, Dong and Xie, Qianqian},
title = {FINCON: a synthesized LLM multi-agent system with conceptual verbal reinforcement for enhanced financial decision making},
year = {2024},
isbn = {9798331314385},
publisher = {Curran Associates Inc.},
address = {Red Hook, NY, USA},
booktitle = {Proceedings of the 38th International Conference on Neural Information Processing Systems},
articleno = {4354},
numpages = {36},
location = {Vancouver, BC, Canada},
series = {NIPS '24}
}

@misc{chen2025stockbench,
      title={StockBench: Can LLM Agents Trade Stocks Profitably In Real-world Markets?}, 
      author={Yanxu Chen and Zijun Yao and Yantao Liu and Amy Xin and Jin Ye and Jianing Yu and Lei Hou and Juanzi Li},
      year={2026},
      eprint={2510.02209},
      archivePrefix={arXiv},
      primaryClass={cs.LG},
      url={https://arxiv.org/abs/2510.02209}, 
}

@misc{xiong2025flagtrader,
      title={FLAG-Trader: Fusion LLM-Agent with Gradient-based Reinforcement Learning for Financial Trading}, 
      author={Guojun Xiong and Zhiyang Deng and Keyi Wang and Yupeng Cao and Haohang Li and Yangyang Yu and Xueqing Peng and Mingquan Lin and Kaleb E Smith and Xiao-Yang Liu and Jimin Huang and Sophia Ananiadou and Qianqian Xie},
      year={2025},
      eprint={2502.11433},
      archivePrefix={arXiv},
      primaryClass={cs.AI},
      url={https://arxiv.org/abs/2502.11433}, 
}

@INPROCEEDINGS{darmanin2025llmguidedrl,
  author={Darmanin, Adam and Vella, Vince},
  booktitle={2025 3rd International Conference on Foundation and Large Language Models (FLLM)}, 
  title={Language Model Guided Reinforcement Learning in Quantitative Trading}, 
  year={2025},
  volume={},
  number={},
  pages={405-412},
  doi={10.1109/FLLM67465.2025.11390898}}

@misc{grover2025finrllama,
      title={FinRLlama: A Solution to LLM-Engineered Signals Challenge at FinRL Contest 2024}, 
      author={Arnav Grover},
      year={2025},
      eprint={2502.01992},
      archivePrefix={arXiv},
      primaryClass={q-fin.TR},
      url={https://arxiv.org/abs/2502.01992}, 
}

@inproceedings{zhang2025finllmb,
    title = "{F}in{LLM}-{B}: When Large Language Models Meet Financial Breakout Trading",
    author = "Zhang, Kang  and
      Yoshie, Osamu  and
      Sun, Lichao  and
      Huang, Weiran",
    editor = "Chen, Weizhu  and
      Yang, Yi  and
      Kachuee, Mohammad  and
      Fu, Xue-Yong",
    booktitle = "Proceedings of the 2025 Conference of the Nations of the Americas Chapter of the Association for Computational Linguistics: Human Language Technologies (Volume 3: Industry Track)",
    month = apr,
    year = "2025",
    address = "Albuquerque, New Mexico",
    publisher = "Association for Computational Linguistics",
    url = "https://aclanthology.org/2025.naacl-industry.29/",
    doi = "10.18653/v1/2025.naacl-industry.29",
    pages = "349--357",
    ISBN = "979-8-89176-194-0"
}

@misc{benhenda2025finrldeepseek,
      title={FinRL-DeepSeek: LLM-Infused Risk-Sensitive Reinforcement Learning for Trading Agents}, 
      author={Mostapha Benhenda},
      year={2025},
      eprint={2502.07393},
      archivePrefix={arXiv},
      primaryClass={q-fin.TR},
      url={https://arxiv.org/abs/2502.07393}, 
}

@article{Wang_Huang_Tu_Zhang_Xu_2021, title={DeepTrader: A Deep Reinforcement Learning Approach for Risk-Return Balanced Portfolio Management with Market Conditions Embedding}, volume={35}, url={https://ojs.aaai.org/index.php/AAAI/article/view/16144}, DOI={10.1609/aaai.v35i1.16144}, abstractNote={Most existing reinforcement learning (RL)-based portfolio management models do not take into account the market conditions, which limits their performance in risk-return balancing. In this paper, we propose DeepTrader, a deep RL method to optimize the investment policy. In particular, to tackle the risk-return balancing problem, our model embeds macro market conditions as an indicator to dynamically adjust the proportion between long and short funds, to lower the risk of market fluctuations, with the negative maximum drawdown as the reward function. Additionally, the model involves a unit to evaluate individual assets, which learns dynamic patterns from historical data with the price rising rate as the reward function. Both temporal and spatial dependencies between assets are captured hierarchically by a specific type of graph structure. Particularly, we find that the estimated causal structure best captures the interrelationships between assets, compared to industry classification and correlation. The two units are complementary and integrated to generate a suitable portfolio which fits the market trend well and strikes a balance between return and risk effectively. Experiments on three well-known stock indexes demonstrate the superiority of DeepTrader in terms of risk-gain criteria.}, number={1}, journal={Proceedings of the AAAI Conference on Artificial Intelligence}, author={Wang, Zhicheng and Huang, Biwei and Tu, Shikui and Zhang, Kun and Xu, Lei}, year={2021}, month={May}, pages={643–650} }

@inproceedings{ijcai2020p641,
  title     = {Relation-Aware Transformer for Portfolio Policy Learning},
  author    = {Xu, Ke and Zhang, Yifan and Ye, Deheng and Zhao, Peilin and Tan, Mingkui},
  booktitle = {Proceedings of the Twenty-Ninth International Joint Conference on
               Artificial Intelligence, {IJCAI-20}},
  publisher = {International Joint Conferences on Artificial Intelligence Organization},
  editor    = {Christian Bessiere},
  pages     = {4647--4653},
  year      = {2020},
  month     = {7},
  note      = {Special Track on AI in FinTech},
  doi       = {10.24963/ijcai.2020/641},
  url       = {https://doi.org/10.24963/ijcai.2020/641},
}

@inproceedings{10.24963/ijcai.2023/548,
author = {Yang, Mengyuan and Zhu, Mengying and Liang, Qianqiao and Zheng, Xiaolin and Wang, MengHan},
title = {Spotlight news driven quantitative trading based on trajectory optimization},
year = {2023},
isbn = {978-1-956792-03-4},
url = {https://doi.org/10.24963/ijcai.2023/548},
doi = {10.24963/ijcai.2023/548},
booktitle = {Proceedings of the Thirty-Second International Joint Conference on Artificial Intelligence},
articleno = {548},
numpages = {10},
location = {Macao, P.R.China},
series = {IJCAI '23}
}

@inproceedings{ijcai2022p557,
  title     = {A Smart Trader for Portfolio Management based on Normalizing Flows},
  author    = {Yang, Mengyuan and Zheng, Xiaolin and Liang, Qianqiao and Han, Bing and Zhu, Mengying},
  booktitle = {Proceedings of the Thirty-First International Joint Conference on
               Artificial Intelligence, {IJCAI-22}},
  publisher = {International Joint Conferences on Artificial Intelligence Organization},
  editor    = {Lud De Raedt},
  pages     = {4014--4021},
  year      = {2022},
  month     = {7},
  note      = {Main Track},
  doi       = {10.24963/ijcai.2022/557},
  url       = {https://doi.org/10.24963/ijcai.2022/557},
}

@InProceedings{pmlr-v235-ma24l,
  title = 	 {Reward Shaping for Reinforcement Learning with An Assistant Reward Agent},
  author =       {Ma, Haozhe and Sima, Kuankuan and Vo, Thanh Vinh and Fu, Di and Leong, Tze-Yun},
  booktitle = 	 {Proceedings of the 41st International Conference on Machine Learning},
  pages = 	 {33925--33939},
  year = 	 {2024},
  editor = 	 {Salakhutdinov, Ruslan and Kolter, Zico and Heller, Katherine and Weller, Adrian and Oliver, Nuria and Scarlett, Jonathan and Berkenkamp, Felix},
  volume = 	 {235},
  series = 	 {Proceedings of Machine Learning Research},
  month = 	 {21--27 Jul},
  publisher =    {PMLR},
  url = 	 {https://proceedings.mlr.press/v235/ma24l.html}
}

@InProceedings{pmlr-v235-wang24bh,
  title = 	 {{LLM}-Empowered State Representation for Reinforcement Learning},
  author =       {Wang, Boyuan and Qu, Yun and Jiang, Yuhang and Shao, Jianzhun and Liu, Chang and Yang, Wenming and Ji, Xiangyang},
  booktitle = 	 {Proceedings of the 41st International Conference on Machine Learning},
  pages = 	 {51348--51375},
  year = 	 {2024},
  editor = 	 {Salakhutdinov, Ruslan and Kolter, Zico and Heller, Katherine and Weller, Adrian and Oliver, Nuria and Scarlett, Jonathan and Berkenkamp, Felix},
  volume = 	 {235},
  series = 	 {Proceedings of Machine Learning Research},
  month = 	 {21--27 Jul},
  publisher =    {PMLR},
  url = 	 {https://proceedings.mlr.press/v235/wang24bh.html}
}

@inproceedings{10.5555/2772879.2772905,
author = {Brys, Tim and Harutyunyan, Anna and Taylor, Matthew E. and Now\'{e}, Ann},
title = {Policy Transfer using Reward Shaping},
year = {2015},
isbn = {9781450334136},
publisher = {International Foundation for Autonomous Agents and Multiagent Systems},
address = {Richland, SC},
booktitle = {Proceedings of the 2015 International Conference on Autonomous Agents and Multiagent Systems},
pages = {181–188},
numpages = {8},
location = {Istanbul, Turkey},
series = {AAMAS '15}
}

@inproceedings{10.1145/3637528.3672064,
author = {Zong, Chuqiao and Wang, Chaojie and Qin, Molei and Feng, Lei and Wang, Xinrun and An, Bo},
title = {MacroHFT: Memory Augmented Context-aware Reinforcement Learning On High Frequency Trading},
year = {2024},
isbn = {9798400704901},
publisher = {Association for Computing Machinery},
address = {New York, NY, USA},
url = {https://doi.org/10.1145/3637528.3672064},
doi = {10.1145/3637528.3672064},
booktitle = {Proceedings of the 30th ACM SIGKDD Conference on Knowledge Discovery and Data Mining},
pages = {4712–4721},
numpages = {10},
location = {Barcelona, Spain},
series = {KDD '24}
}

@inproceedings{ijcai2023p441,
  title     = {Contrastive Learning and Reward Smoothing for Deep Portfolio Management},
  author    = {Lien, Yun-Hsuan and Li, Yuan-Kui and Wang, Yu-Shuen},
  booktitle = {Proceedings of the Thirty-Second International Joint Conference on
               Artificial Intelligence, {IJCAI-23}},
  publisher = {International Joint Conferences on Artificial Intelligence Organization},
  editor    = {Edith Elkind},
  pages     = {3966--3974},
  year      = {2023},
  month     = {8},
  note      = {Main Track},
  doi       = {10.24963/ijcai.2023/441},
  url       = {https://doi.org/10.24963/ijcai.2023/441},
}

@inproceedings{ijcai2020p623,
  title     = {MAPS: Multi-Agent reinforcement learning-based Portfolio management System.},
  author    = {Lee, Jinho and Kim, Raehyun and Yi, Seok-Won and Kang, Jaewoo},
  booktitle = {Proceedings of the Twenty-Ninth International Joint Conference on
               Artificial Intelligence, {IJCAI-20}},
  publisher = {International Joint Conferences on Artificial Intelligence Organization},
  editor    = {Christian Bessiere},
  pages     = {4520--4526},
  year      = {2020},
  month     = {7},
  note      = {Special Track on AI in FinTech},
  doi       = {10.24963/ijcai.2020/623},
  url       = {https://doi.org/10.24963/ijcai.2020/623},
}

@inproceedings{10.1145/3580305.3599315,
author = {Zhao, Lifan and Kong, Shuming and Shen, Yanyan},
title = {DoubleAdapt: A Meta-learning Approach to Incremental Learning for Stock Trend Forecasting},
year = {2023},
isbn = {9798400701030},
publisher = {Association for Computing Machinery},
address = {New York, NY, USA},
url = {https://doi.org/10.1145/3580305.3599315},
doi = {10.1145/3580305.3599315},
booktitle = {Proceedings of the 29th ACM SIGKDD Conference on Knowledge Discovery and Data Mining},
pages = {3492–3503},
numpages = {12},
location = {Long Beach, CA, USA},
series = {KDD '23}
}

@article{Ma2023EurekaHR,
  title={Eureka: Human-Level Reward Design via Coding Large Language Models},
  author={Yecheng Jason Ma and William Liang and Guanzhi Wang and De-An Huang and Osbert Bastani and Dinesh Jayaraman and Yuke Zhu and Linxi Fan and Anima Anandkumar},
  journal={ArXiv},
  year={2023},
  volume={abs/2310.12931},
  url={https://api.semanticscholar.org/CorpusID:264306288}
}

@INPROCEEDINGS{10658032,
  author={Li, Hao and Yang, Xue and Wang, Zhaokai and Zhu, Xizhou and Zhou, Jie and Qiao, Yu and Wang, Xiaogang and Li, Hongsheng and Lu, Lewei and Dai, Jifeng},
  booktitle={2024 IEEE/CVF Conference on Computer Vision and Pattern Recognition (CVPR)}, 
  title={Auto MC-Reward: Automated Dense Reward Design with Large Language Models for Minecraft}, 
  year={2024},
  volume={},
  number={},
  pages={16426-16435},
  doi={10.1109/CVPR52733.2024.01554}}

@book{bollinger2002bollinger,
  title = {Bollinger on Bollinger Bands},
  author = {Bollinger, J.},
  year = 2002,
  publisher = {McGraw-Hill Education},
  isbn = {978-0-07-137368-5},
  lccn = {2001030666}
}

@inproceedings{chen2016xgboost,
author = {Chen, Tianqi and Guestrin, Carlos},
title = {XGBoost: A Scalable Tree Boosting System},
year = {2016},
isbn = {9781450342322},
publisher = {Association for Computing Machinery},
address = {New York, NY, USA},
url = {https://doi.org/10.1145/2939672.2939785},
doi = {10.1145/2939672.2939785},
booktitle = {Proceedings of the 22nd ACM SIGKDD International Conference on Knowledge Discovery and Data Mining},
pages = {785–794},
numpages = {10},
location = {San Francisco, California, USA},
series = {KDD '16}
}

@article{mcconnell2008equity,
  title={Equity returns at the turn of the month},
  author={McConnell, John J and Xu, Wei},
  journal={Financial Analysts Journal},
  volume={64},
  number={2},
  pages={49--64},
  year={2008},
  publisher={Taylor \& Francis}
}

\clearpage
\appendix

\section{Related Work}
\label{app:related_work}

\paragraph{DRL for portfolio trading.}
Portfolio trading is naturally formulated as a sequential decision-making problem, where an agent observes market conditions, chooses rebalancing actions, and receives feedback through returns, risk exposure, and trading frictions.
Unlike two-stage pipelines that first predict returns and then derive trading signals, reinforcement learning directly optimizes dynamic portfolio decisions and has become an important approach for quantitative trading and portfolio management~\cite{moody2001direct, jiang2017portfolio,sun2023rlqt,hambly2023rlfinance}.
Recent surveys and benchmarks further summarize portfolio DRL methods, including value-based methods and actor--critic algorithms such as PPO, A2C, DDPG, SAC, and TD3~\cite{pippas2025rlqfsurvey,rezaei2025drlpm,liu2022finrlmeta,schulman2017ppo}.

Most existing work focuses on policy architectures, optimization algorithms, or market simulators, while treating the state representation and reward function as fixed interfaces.
The agent is often trained on predefined OHLCV variables, technical indicators, or static financial features, and the reward is usually based on short-horizon return or manually designed risk penalties.
GIFT is complementary to this line of work.
Rather than replacing downstream learners such as PPO or A2C, it uses an LLM to design and refine the state-reward interface used by the RL agent.

\paragraph{State-reward interface design in financial RL.}
State and reward design determine what market structure the agent can observe and which behaviors are reinforced during training.
Prior work has improved trading policies by adding positional features, technical indicators, data fusion, and robust representations to capture information beyond raw price-volume inputs~\cite{goluza2024positional,liang2018adversarial}.
On the reward side, profit or short-horizon return can provide noisy feedback in non-stationary markets, especially under transaction costs, drawdown risk, and low signal-to-noise conditions.
Reward shaping, risk-sensitive rewards, transaction-cost-aware objectives, regret-based Sharpe rewards, and imitation rewards have therefore been studied to provide more stable and risk-aware learning signals~\cite{ng1999rewardshaping,goluza2024imitationreward,karzanov2025regretreward}.

These studies show that states and rewards are central learning interfaces in financial RL, yet most existing designs are manually specified before training and are not revised using training dynamics, market windows, or portfolio behavior.
GIFT addresses this limitation through LLM-guided offline interface design, where financial factors, risk rules, and rollout diagnostics constrain executable state-enhancement and reward-shaping candidates before evaluation.

\paragraph{LLMs for financial decision making.}
Financial LLMs and related benchmarks have been studied for financial text understanding, sentiment analysis, risk management, reasoning, and decision support, with performance depending heavily on domain corpora, task formulation, and evaluation protocols~\cite{wu2023bloomberggpt,yang2023fingpt,nie2024finllmsurvey,xie2024finben}.
For trading, recent work explores multi-agent investment systems, LLM-based policy networks, LLM-generated trading signals, and LLM--RL hybrids with market feedback or policy-gradient training~\cite{yu2024fincon,xiong2025flagtrader,darmanin2025llmguidedrl,grover2025finrllama}.

These studies suggest that LLMs can organize financial information and provide high-level cues, but recent benchmarks also report risks such as look-ahead bias, data contamination, hallucinated rationales, inconsistent reasoning, insufficient risk control, and unstable out-of-sample behavior~\cite{chen2025stockbench,zhang2025finllmb,benhenda2025finrldeepseek}.

GIFT therefore does not use the LLM as a trading agent, policy network, or portfolio-weight generator.
Instead, it restricts the LLM to the state-reward design layer: given financial factors, risk rules, and rollout diagnostics, the LLM proposes executable interface candidates, while final actions are learned by the downstream DRL policy.
This controlled role uses LLMs to structure financial knowledge without exposing trading decisions to open-ended language generation.

\section{Preliminary Evidence for Feedback-Guided Interface Refinement}
\label{app:preliminary_feedback}

This section details the diagnostic experiment in Figure~\ref{fig:motivation}. 
The goal is not to evaluate final portfolio performance, but to examine whether different state-reward interfaces produce different PPO training-feedback patterns. 
This connects to the motivation in the introduction: raw OHLCV states may leave useful financial structure implicit, while short-horizon return rewards can be noisy and weakly aligned with long-term return-risk objectives. 
If financial guidance improves the interface, PPO should receive reward trajectories that are more directional, temporally coherent, and less dominated by episode-level fluctuations.

\paragraph{Compared settings.}
We compare three PPO settings under the same backbone and training protocol. 
\textbf{Vanilla PPO} uses the original state and reward. 
\textbf{Free-form LLM-PPO} uses LLM-generated state and reward designs without financial-factor or risk-rule constraints. 
\textbf{Guided LLM-PPO} constrains LLM generation with the proposed financial factor and risk-rule libraries. 
Thus, the comparison separates the effect of financial guidance from the mere use of an LLM.

\paragraph{Diagnostics.}
For each rolling window, let \(\{r_e\}_{e=1}^{E}\) be the episode-level PPO training reward sequence over \(E=50\) episodes, where \(r_e\) is the cumulative reward in episode \(e\). 
We compute four complementary diagnostics:
\[
\begin{aligned}
\Delta_{50} &= r_{50}-r_1, \\
\beta &= \mathrm{Cov}(e,r_e)/\mathrm{Var}(e), \\
\rho_1 &= \mathrm{corr}(r_{1:E-1}, r_{2:E}), \\
\mathrm{SNR} &= |\beta|/(\mathrm{Std}(\Delta r_e)+\varepsilon),
\quad \Delta r_e=r_{e+1}-r_e .
\end{aligned}
\]
Here, \(\Delta_{50}\) measures cumulative training improvement, \(\beta\) captures the direction of the reward trend, \(\rho_1\) measures temporal coherence, and SNR compares the trend strength with short-term reward fluctuations. 
Together, these metrics characterize whether the reward sequence provides a cleaner learning signal, rather than only whether the final episode reward is higher.

\begin{table}[t]
\centering
\scriptsize
\setlength{\tabcolsep}{4.5pt}
\renewcommand{\arraystretch}{1.12}
\caption{
Training-feedback diagnostics in the preliminary experiment.
Higher values indicate clearer PPO training feedback.
}
\label{tab:preliminary_feedback_summary}
\resizebox{\columnwidth}{!}{
\begin{tabular}{lcccc}
\toprule
\textbf{Setting}
& \(\boldsymbol{\Delta_{50}}\uparrow\)
& \(\boldsymbol{\beta}\uparrow\)
& \(\boldsymbol{\rho_1}\uparrow\)
& \textbf{SNR}\(\uparrow\) \\
\midrule
Vanilla PPO
& $-0.49$
& $+0.003$
& $0.051$
& $0.013$ \\
Free-form LLM-PPO
& $-0.08$
& $-0.015$
& $-0.031$
& $0.007$ \\
Guided LLM-PPO
& $\mathbf{+19.42}$
& $\mathbf{+0.313}$
& $\mathbf{0.797}$
& $\mathbf{0.149}$ \\
\bottomrule
\end{tabular}
}
\end{table}

\paragraph{Results.}
Table~\ref{tab:preliminary_feedback_summary} shows that Guided LLM-PPO obtains the strongest values on all four diagnostics. 
The positive \(\Delta_{50}\) and larger \(\beta\) indicate clearer training improvement and a stronger upward reward trend. 
The higher \(\rho_1\) suggests that adjacent episode rewards are more coherent, while the higher SNR indicates that the trend is less obscured by episode-level fluctuations.

Free-form LLM-PPO does not show the same improvement, despite also using LLM-generated designs. 
Its reward trend and autocorrelation are negative on average, and its SNR is lower than Vanilla PPO. 
This suggests that open-ended LLM generation can introduce unstable or weakly relevant signals if it is not constrained by financial factors and risk rules.

Overall, these diagnostics support the motivation for GIFT's constrained interface-design strategy: financial guidance helps restrict LLM-generated state-reward designs to a more structured search space, where PPO receives more directional and coherent training feedback. 
We use this result only as training-dynamics evidence; out-of-sample portfolio performance is evaluated separately in the main experiments.

\section{Details of Methods}

\subsection{Offline Interface Design Algorithm}
\label{app:offline_interface_algorithm}

Algorithm~\ref{alg:gift} summarizes the offline interface-design procedure of GIFT.
For each rolling window, GIFT uses only the historical design window to generate and refine executable state-reward interfaces.
Each candidate interface consists of a factor-based state-enhancement function \(F\), a shaped reward function \(R\), and the selected risk-rule terms.
Valid candidates are trained with PPO on the design window and ranked by rollout-level feedback.
The selected interface is then frozen before evaluation, so test-time policy learning uses the same fixed interface without further LLM queries or interface updates.

\begin{algorithm}[htbp]
\footnotesize
\caption{GIFT: Offline Financial-Knowledge-Guided Interface Design}
\label{alg:gift}
\begin{algorithmic}[1]
\REQUIRE Factor library \(\mathcal{B}\), risk-rule library \(\mathcal{R}\), statistics \(\mathrm{Stats}\), rounds \(M\), candidates \(K\)
\ENSURE Frozen state--reward interface \((F^\star,R^\star)\)

\STATE Initialize feedback memory \(\mathcal{H}_1\leftarrow\emptyset\) and candidate pool \(\mathcal{G}\leftarrow\emptyset\)
\FOR{\(m=1,\dots,M\)}
    \STATE \(\mathcal{C}_m \leftarrow \mathcal{M}_{\mathrm{LLM}}(\mathcal{B},\mathcal{R},\mathrm{Stats}_m,\mathcal{H}_m,K)\)
    \STATE \(\widetilde{\mathcal{C}}_m \leftarrow \mathrm{Validate}(\mathcal{C}_m)\)
    \FORALL{\((F_{m,k},R_{m,k})\in\widetilde{\mathcal{C}}_m\)}
        \STATE Train PPO with \((F_{m,k},R_{m,k})\)
        \STATE Compute score \(S^{\mathrm{ro}}_{m,k}\) and diagnostics \(\mathcal{D}_{m,k}\)
        \STATE \(\mathcal{G}\leftarrow \mathcal{G}\cup\{(F_{m,k},R_{m,k},S^{\mathrm{ro}}_{m,k},\mathcal{D}_{m,k})\}\)
    \ENDFOR
    \STATE \(k_m^\star \leftarrow \arg\max_{k:(F_{m,k},R_{m,k})\in\widetilde{\mathcal{C}}_m} S^{\mathrm{ro}}_{m,k}\)
    \STATE \(\mathcal{H}_{m+1}\leftarrow \mathrm{Update}(\mathcal{H}_m,\mathcal{D}_{m,k_m^\star})\)
\ENDFOR
\STATE \((F^\star,R^\star)\leftarrow \arg\max_{(F,R,S^{\mathrm{ro}},\mathcal{D})\in\mathcal{G}} S^{\mathrm{ro}}\)
\STATE Freeze \((F^\star,R^\star)\); no LLM query, prompt update, feedback write, or interface modification during evaluation
\RETURN \((F^\star,R^\star)\)
\end{algorithmic}
\end{algorithm}

\subsection{Factor and Risk-Rule Libraries for FSE and RRS}
\label{app:library}

This section describes the two constrained libraries used by GIFT: the financial factor library \(\mathcal{B}\) for Factor-guided State Enhancement (FSE), and the risk-rule library \(\mathcal{R}\) for Risk-rule-guided Reward Shaping (RRS). 
The LLM can select, parameterize, and compose primitives from these libraries, but cannot introduce arbitrary features or reward terms. 
This keeps the generated interface executable, bounded, and easier to inspect.

\subsection{Financial Factor Library for FSE}

\paragraph{\bfseries\itshape Factor construction.}
The factor library is implemented in Python and NumPy without external technical-analysis packages such as \texttt{ta-lib} or \texttt{pandas\_ta}. 
Each factor has default parameters and valid ranges; LLM-specified parameters are clipped before execution. 
The generated state-enhancement function maps a raw look-back window to a normalized feature vector, with numerical guards for NaN and Inf values.

The raw state uses a \(W=20\)-day look-back window. 
Each day contains six channels, \texttt{[close, open, high, low, volume, adj\_close]}, forming a 120-dimensional interleaved vector. 
The library includes 20 registered indicators and 9 reusable building-block functions.

\paragraph{\bfseries\itshape Registered indicators.}
Table~\ref{tab:factor_library_indicators} lists the registered indicators in \(\mathcal{B}\). 
Although ADX is grouped with volume-related indicators in implementation, we treat it as a trend-strength signal when describing the financial coverage of \(\mathcal{B}\).

\begin{table*}[t]
\centering
\scriptsize
\setlength{\tabcolsep}{3.2pt}
\renewcommand{\arraystretch}{1.08}
\caption{
Registered indicators in the financial factor library \(\mathcal{B}\).
}
\label{tab:factor_library_indicators}
\resizebox{\textwidth}{!}{
\begin{tabular}{lllll}
\toprule
\textbf{Theme} & \textbf{Indicator} & \textbf{Dim.} & \textbf{Default parameters} & \textbf{Valid parameter ranges} \\
\midrule
Trend & RSI & 1 & window=14 & window \(\in[5,60]\) \\
Trend & MACD & 3 & fast=12, slow=26, signal=9 & fast \(\in[5,20]\), slow \(\in[15,60]\), signal \(\in[3,15]\) \\
Trend & EMA\_Cross & 1 & fast=12, slow=26 & fast \(\in[5,20]\), slow \(\in[15,60]\) \\
Trend & Momentum & 1 & window=10 & window \(\in[5,60]\) \\
Trend & ROC & 1 & window=10 & window \(\in[5,60]\) \\
Trend & SMA\_Cross & 1 & fast=10, slow=30 & fast \(\in[5,20]\), slow \(\in[15,60]\) \\
Trend & DEMA & 1 & window=20 & window \(\in[5,60]\) \\
Trend & TSF & 1 & window=14 & window \(\in[5,30]\) \\
Trend & Williams\_Alligator & 3 & fixed windows & -- \\
\midrule
Volatility & Bollinger & 3 & window=20, num\_std=2.0 & window \(\in[10,40]\), num\_std \(\in[1.0,3.0]\) \\
Volatility & ATR & 1 & window=14 & window \(\in[5,30]\) \\
Volatility & Volatility & 1 & window=20 & window \(\in[5,60]\) \\
Volatility & Skewness & 1 & window=20 & window \(\in[5,60]\) \\
Volatility & Kurtosis & 1 & window=20 & window \(\in[5,60]\) \\
\midrule
Mean reversion & Stochastic & 2 & window=14 & window \(\in[5,30]\) \\
Mean reversion & Williams\_R & 1 & window=14 & window \(\in[5,30]\) \\
Mean reversion & CCI & 1 & window=20 & window \(\in[5,30]\) \\
\midrule
Volume & OBV & 1 & -- & -- \\
Volume & Volume\_Ratio & 1 & window=20 & window \(\in[5,30]\) \\
Volume / trend strength & ADX & 1 & window=14 & window \(\in[5,30]\) \\
\bottomrule
\end{tabular}
}
\end{table*}

\paragraph{\bfseries\itshape Coverage of financial dimensions.}
The library covers six financial dimensions used by FSE.

\paragraph{\bfseries\itshape Building-block functions.}
Table~\ref{tab:factor_library_blocks} lists additional primitives for relative momentum, cross-sectional ranking, realized volatility, downside risk, beta, multi-horizon momentum, mean reversion, and liquidity.

\begin{tcolorbox}[
    colback=gray!3,
    colframe=gray!45,
    boxrule=0.45pt,
    arc=2pt,
    left=5pt,
    right=5pt,
    top=4pt,
    bottom=4pt,
    title={Financial-Factor Coverage},
    fonttitle=\bfseries\small
]
\footnotesize
\RaggedRight
\begin{description}[leftmargin=1.7em, labelsep=0.45em, itemsep=2pt, topsep=0pt]
    \item[\textbf{Momentum}] Momentum, ROC, \fcode{compute_relative_momentum}, \fcode{compute_multi_horizon_momentum}.
    \item[\textbf{Volatility}] Volatility, ATR, Bollinger, \fcode{compute_realized_volatility}.
    \item[\textbf{Downside risk}] \fcode{compute_downside_risk}, Skewness, Kurtosis, \fcode{compute_beta}.
    \item[\textbf{Liquidity}] OBV, Volume\_Ratio, \fcode{compute_turnover_ratio}.
    \item[\textbf{Mean reversion}] Stochastic, Williams\_R, CCI, \fcode{compute_zscore_price}, \fcode{compute_mean_reversion_signal}.
    \item[\textbf{Trend strength}] ADX, RSI, MACD, EMA\_Cross, SMA\_Cross, DEMA, TSF, Williams\_Alligator.
\end{description}
\end{tcolorbox}

\begin{table*}[t]
\centering
\scriptsize
\setlength{\tabcolsep}{3.5pt}
\renewcommand{\arraystretch}{1.08}
\caption{
Building-block functions in the financial factor library \(\mathcal{B}\).
}
\label{tab:factor_library_blocks}
\resizebox{\textwidth}{!}{
\begin{tabular}{llll}
\toprule
\textbf{Function} & \textbf{Main input} & \textbf{Dim.} & \textbf{Description} \\
\midrule
\texttt{compute\_relative\_momentum} & prices, window=20 & 1 & Relative price momentum over the look-back window \\
\texttt{compute\_cross\_sectional\_rank} & values & 1 & Cross-sectional rank normalized to \([0,1]\) \\
\texttt{compute\_realized\_volatility} & returns, window=20 & 1 & Realized volatility based on return standard deviation \\
\texttt{compute\_downside\_risk} & returns, window=20 & 1 & Downside semi-deviation using negative returns only \\
\texttt{compute\_beta} & returns, market\_returns, window=20 & 1 & Rolling beta relative to the equal-weighted portfolio \\
\texttt{compute\_multi\_horizon\_momentum} & prices, windows=[5,10,20] & 3 & Short-, medium-, and long-horizon momentum \\
\texttt{compute\_zscore\_price} & prices, window=20 & 1 & Price z-score relative to the moving average \\
\texttt{compute\_mean\_reversion\_signal} & prices, window=20 & 1 & Mean-reversion strength based on z-score deviation \\
\texttt{compute\_turnover\_ratio} & volumes, window=20 & 1 & Current volume relative to average volume \\
\bottomrule
\end{tabular}
}
\end{table*}

\subsection{Risk-Rule Library for RRS}
\paragraph{\bfseries\itshape Risk-rule construction.}
The risk-rule library contains seven portfolio-level reward-shaping rules. 
Each rule uses portfolio context such as current weights, previous weights, market state, and drawdown. 
As with the factor library, each rule has default parameters and valid ranges, and LLM-specified parameters are clipped before execution.

At each refinement round, the LLM selects and parameterizes a compact subset of 2--4 rules. 
The selected rules are assembled into a reward-shaping function added to the base portfolio reward. 
In our setting, the first five dimensions of the action vector correspond to risky assets, and the last dimension corresponds to cash.

\paragraph{\bfseries\itshape Registered risk rules.}
Table~\ref{tab:risk_rule_library} lists the registered rules in \(\mathcal{R}\).

\begin{table*}[t]
\centering
\scriptsize
\setlength{\tabcolsep}{3.5pt}
\renewcommand{\arraystretch}{1.08}
\caption{
Registered rules in the risk-rule library \(\mathcal{R}\).
}
\label{tab:risk_rule_library}
\resizebox{\textwidth}{!}{
\begin{tabular}{llll}
\toprule
\textbf{Rule} & \textbf{Default parameters} & \textbf{Valid parameter ranges} & \textbf{Effect} \\
\midrule
\texttt{penalize\_concentration}
& max\_weight=0.35, penalty=0.1
& max\_weight \(\in[0.2,0.5]\), penalty \(\in[0.01,0.2]\)
& Penalizes stock weights above max\_weight \\

\texttt{reward\_diversification}
& min\_stocks=3, bonus=0.05
& min\_stocks \(\in[2,5]\), bonus \(\in[0.01,0.1]\)
& Rewards holding at least min\_stocks with weights above 5\% \\

\texttt{penalize\_turnover}
& threshold=0.1, penalty=0.15
& threshold \(\in[0.05,0.3]\), penalty \(\in[0.01,0.2]\)
& Penalizes turnover above threshold \\

\texttt{regime\_defensive}
& crisis\_threshold=0.6, cash\_bonus=0.1
& crisis\_threshold \(\in[0.5,0.8]\), cash\_bonus \(\in[0.01,0.15]\)
& Rewards cash allocation in high-risk regimes \\

\texttt{momentum\_alignment}
& bonus=0.05
& bonus \(\in[0.01,0.1]\)
& Rewards positive alignment between weights and momentum ranks \\

\texttt{volatility\_scaling}
& vol\_threshold=0.5, scale=0.5
& vol\_threshold \(\in[0.3,0.8]\), scale \(\in[0.3,0.8]\)
& Downscales the base reward in high-volatility states \\

\texttt{drawdown\_penalty}
& dd\_threshold=0.1, penalty=0.15
& dd\_threshold \(\in[0.05,0.2]\), penalty \(\in[0.05,0.3]\)
& Penalizes drawdown above threshold \\
\bottomrule
\end{tabular}
}
\end{table*}

\paragraph{\bfseries\itshape Coverage of risk-control dimensions.}
The risk-rule library \(\mathcal{R}\) covers six risk-control dimensions:

\begin{itemize}[leftmargin=*, itemsep=1pt, topsep=2pt]
    \item \textbf{Concentration/diversification:} \texttt{penalize\_concentration}, \texttt{reward\_diversification}.
    \item \textbf{Turnover:} \texttt{penalize\_turnover}.
    \item \textbf{Drawdown:} \texttt{drawdown\_penalty}.
    \item \textbf{Volatility:} \texttt{volatility\_scaling}.
    \item \textbf{Regime defense:} \texttt{regime\_defensive}.
    \item \textbf{Momentum alignment:} \texttt{momentum\_alignment}.
\end{itemize}

\definecolor{GIFTkw}{RGB}{0,0,180}
\definecolor{GIFTstr}{RGB}{163,21,21}
\definecolor{GIFTcom}{RGB}{0,128,0}
\definecolor{GIFTbg}{RGB}{248,248,248}
\definecolor{GIFTrule}{RGB}{200,200,200}

\lstdefinestyle{GIFTplain}{
  basicstyle=\ttfamily\scriptsize,
  backgroundcolor=\color{GIFTbg},
  frame=single,
  rulecolor=\color{GIFTrule},
  breaklines=true,
  columns=fullflexible,
  keepspaces=true,
  showstringspaces=false,
  xleftmargin=4pt,
  xrightmargin=2pt,
}

\lstdefinestyle{GIFTpy}{
  style=GIFTplain,
  language=Python,
  keywordstyle=\color{GIFTkw}\bfseries,
  stringstyle=\color{GIFTstr},
  commentstyle=\color{GIFTcom}\itshape,
  morekeywords={np},
}

\lstdefinestyle{GIFTjson}{
  style=GIFTplain,
  stringstyle=\color{GIFTstr},
  morestring=[b]",
}
% --------------------------------------------------------------------

\section{Prompt Details of Our Methods}

\subsection{Prompt Templates and Generation Protocol}
\label{app:prompt_details}

This section reports the prompt templates used by GIFT to generate and refine the state-reward interface. 
GIFT uses three prompt families: Factor-guided State Enhancement (FSE; Appendix~\ref{app:prompt_fse}), Risk-rule-guided Reward Shaping (RRS; Appendix~\ref{app:prompt_rrs}), and Diagnostic-guided Refinement (DGR; Appendix~\ref{app:prompt_dgr}). 
All prompts use the same design principle: the LLM proposes executable interface components, while PPO remains responsible for learning portfolio actions.

% ---------------------------------------------------------------------
\subsubsection{Prompting Workflow for Interface Generation}
\label{app:prompt_overview}

Within each training window, GIFT searches for a state-reward interface through three LLM-assisted steps.

\begin{tcolorbox}[
    colback=gray!3,
    colframe=gray!45,
    boxrule=0.45pt,
    arc=2pt,
    left=5pt,
    right=5pt,
    top=4pt,
    bottom=4pt,
    title={Prompting Workflow},
    fonttitle=\bfseries\small
]
\footnotesize
\raggedright

\noindent\textbf{1. State and intrinsic-reward generation.}
Generate \promptcode{revise\_state(s)} and \promptcode{intrinsic\_reward(updated\_s)} to append factor channels and provide dense intrinsic feedback.

\vspace{3pt}
\noindent\textbf{2. Risk-rule selection.}
Select a compact set of portfolio-level risk rules and set the drawdown penalty coefficient.

\vspace{3pt}
\noindent\textbf{3. Diagnostic-guided refinement.}
Use PPO rollout diagnostics to revise the previous interface in the next round.
\end{tcolorbox}

The LLM is used only at the interface-design stage. 
It never outputs portfolio weights or trading actions. 
After search, the selected interface is fixed before out-of-sample evaluation.

% ---------------------------------------------------------------------
\subsubsection{Shared Context for Module-Specific Prompts}
\label{app:shared_context}

All prompt families share the following context.

\paragraph{\bfseries\itshape Input state layout.}
For each equity, the raw state is a 120-dimensional vector containing a 20-day history with six market channels per day.

\begin{minipage}{0.95\linewidth}
\begin{lstlisting}[style=GIFTplain]
s[0..5]     = [close, open, high, low, volume, adj_close] for day 1
s[6..11]    = same six channels                           for day 2
...
s[114..119] = same six channels                           for day 20

Channel slices:
s[0::6] = close      s[1::6] = open      s[2::6] = high
s[3::6] = low        s[4::6] = volume    s[5::6] = adjusted close
\end{lstlisting}
\end{minipage}

The portfolio contains five equities and cash. 
Actions are long-only portfolio weights on the simplex.

\paragraph{\bfseries\itshape Bounded generation space.}
FSE can only construct additional state channels from bounded financial factor types, including momentum, multi-horizon momentum, realized volatility, downside risk, price z-score, mean reversion, and volume-based liquidity. 
Portfolio-level or cross-sectional terms are not generated inside \promptcode{revise\_state(s)}, since they require information beyond a single asset state.

\paragraph{\bfseries\itshape Market-context conditioning.}
Each training window is summarized by return statistics, volatility statistics, asset correlations, and a regime label from 
\{\textsc{Crisis}, \textsc{Defensive}, \textsc{Balanced}, \textsc{Aggressive}\}. 
Crisis and Defensive regimes emphasize risk-sensitive factors and drawdown control; Balanced regimes combine trend, risk, and volume signals; Aggressive regimes allow more trend and momentum features, with bounded risk terms. 
When volatility or asset correlation is high, the prompt additionally emphasizes defensive features, diversification, and turnover control.

\paragraph{\bfseries\itshape Diagnostic feedback.}
After PPO training, appended feature channels are evaluated by predictive correlation with forward returns, critic reliance, and regime-specific correlation. 
The diagnostic summary is used only to revise the next interface candidate.

\begin{tcolorbox}[
    colback=gray!3,
    colframe=gray!45,
    boxrule=0.45pt,
    arc=2pt,
    left=5pt,
    right=5pt,
    top=4pt,
    bottom=4pt,
    title={Diagnostic Feedback Rules},
    fonttitle=\bfseries\small
]
\footnotesize
\raggedright
\textbf{Predictive + used by critic:} retain or extend. \\
\textbf{Used by critic but weakly predictive:} remove or reduce. \\
\textbf{Predictive but underused:} make more explicit. \\
\textbf{Negatively aligned:} remove or invert. \\
\textbf{Weak risk sensitivity:} add volatility or downside-risk terms.
\end{tcolorbox}

% ---------------------------------------------------------------------
\subsubsection{FSE Prompt for Factor-Guided State Enhancement}
\label{app:prompt_fse}

FSE asks the LLM to append compact factor channels to the raw OHLCV state. 
The original 120-dimensional state must be preserved.

\begin{quantprompt}{FSE Prompt --- State Enhancement}
    \scriptsize
    \linespread{0.85}\selectfont
    \color{quanttext}

    \promptlabel{Role:}
    \promptcontent{Act as a quantitative feature designer for portfolio reinforcement learning. Design state features only; do not output actions or portfolio weights.}\\[-1pt]

    \promptlabel{Task:}
    \promptcontent{Generate executable Python code for}
    \promptcode{revise\_state(s)}
    \promptcontent{. Preserve the original 120-dimensional state and append a small set of normalized financial factor channels.}\\[-1pt]

    \promptlabel{Input:}
    \promptcontent{Each equity has a 20-day, six-channel OHLCV-related history. The output must satisfy}
    \promptcode{updated\_s[:120] == s}
    \promptcontent{.}\\[-1pt]

    \promptlabel{Allowed Factors:}
    \promptcontent{Use only bounded factor types such as momentum, multi-horizon momentum, realized volatility, downside risk, price z-score, mean reversion, and volume-based liquidity.}\\[-1pt]

    \promptlabel{Market Context:}
    \promptcontent{Use the provided window-level statistics and regime label to decide which factor types are most relevant.}\\[-1pt]

    \promptlabel{Constraints:}
    \promptcontent{Append only a compact number of features; do not encode portfolio weights, actions, or cash-specific indicators; guard against NaN and Inf values.}\\[-1pt]

    \promptlabel{Output:}
    \promptcontent{Return executable Python code defining}
    \promptcode{revise\_state(s)}
    \promptcontent{and a concise rationale.}
\end{quantprompt}

\begin{minipage}{\linewidth}
\begin{lstlisting}[style=GIFTpy]
import numpy as np

def revise_state(s):
    # Preserve the raw 120-dimensional state.
    # Append normalized financial factor channels.
    return updated_s
\end{lstlisting}
\end{minipage}

% ---------------------------------------------------------------------
\subsubsection{RRS Prompt for Risk-Rule-Guided Reward Shaping}
\label{app:prompt_rrs}

RRS has two parts. 
Part A generates an intrinsic reward over the enhanced state. 
Part B selects portfolio-level risk rules and sets the drawdown penalty coefficient.

\paragraph{\bfseries\itshape Part A: intrinsic reward.}
The intrinsic reward provides additional feedback beyond one-step portfolio return. 
It must use at least one appended factor channel and remain bounded.

\begin{quantprompt}{RRS Prompt --- Intrinsic Reward}
    \scriptsize
    \linespread{0.90}\selectfont
    \color{quanttext}

    \promptlabel{Role:}
    \promptcontent{Design a market-context-conditioned intrinsic reward for PPO portfolio learning. Do not output actions or portfolio weights.}\\[-1pt]

    \promptlabel{Reward Form:}
    \promptcontent{The training reward combines base portfolio return, selected risk rules, and}
    \promptcode{intrinsic\_reward(updated\_s)}
    \promptcontent{.}\\[-1pt]

    \promptlabel{Input:}
    \promptcontent{The function receives}
    \promptcode{updated\_s = revise\_state(s)}
    \promptcontent{and must use at least one appended feature channel with index \(\ge 120\).}\\[-1pt]

    \promptlabel{Design Goal:}
    \promptcontent{Encourage informative market states without duplicating realized return. In defensive regimes, emphasize volatility or downside-risk signals; in balanced or aggressive regimes, allow trend or momentum signals with bounded risk terms.}\\[-1pt]

    \promptlabel{Constraints:}
    \promptcontent{Use simple, interpretable, and numerically stable terms. Keep the reward scale controlled. Return a scalar.}\\[-1pt]

    \promptlabel{Output:}
    \promptcontent{Return executable Python code defining}
    \promptcode{intrinsic\_reward(updated\_s)}
    \promptcontent{and a concise rationale.}
\end{quantprompt}

\paragraph{\bfseries\itshape Part B: risk-rule selection.}
The second RRS prompt keeps reward shaping within a bounded portfolio-level rule space.

\begin{quantprompt}{RRS Prompt --- Risk-Rule Selection}
    \scriptsize
    \linespread{0.90}\selectfont
    \color{quanttext}

    \promptlabel{Role:}
    \promptcontent{Configure portfolio-level reward-shaping rules for PPO.}\\[-1pt]

    \promptlabel{Rule Space:}
    \promptcontent{Choose from concentration penalty, diversification bonus, turnover penalty, defensive cash bonus, momentum alignment, volatility scaling, and drawdown penalty.}\\[-1pt]

    \promptlabel{Selection Policy:}
    \promptcontent{Select 2--4 complementary rules. Defensive regimes should prioritize drawdown, volatility, and cash-holding rules; balanced regimes should mix concentration and turnover control; aggressive regimes should avoid excessive conservatism while retaining basic risk control.}\\[-1pt]

    \promptlabel{Output:}
    \promptcontent{Return valid JSON with selected rules, parameters, drawdown coefficient \(\lambda\), and a brief rationale.}
\end{quantprompt}

\begin{minipage}{\linewidth}
\begin{lstlisting}[style=GIFTjson]
{
  "reward_rules": [
    {"rule": "penalize_concentration", "params": {"max_weight": 0.35}},
    {"rule": "drawdown_penalty", "params": {"threshold": 0.10}}
  ],
  "lambda": 0.6,
  "rationale": "The rules control concentration and drawdown risk."
}
\end{lstlisting}
\end{minipage}

% ---------------------------------------------------------------------
\subsubsection{DGR Prompt for Diagnostic-Guided Refinement}
\label{app:prompt_dgr}

DGR uses PPO rollout diagnostics to revise the previously generated state-reward interface. 
Rather than treating the LLM output as a one-shot design, GIFT feeds back evidence about feature usefulness, reward stability, and portfolio behavior after PPO training. 
The LLM is then asked to make targeted edits, such as retaining predictive features, removing unstable reward terms, adjusting risk rules, or rescaling the drawdown penalty, instead of regenerating the entire interface from scratch. 
% This keeps refinement grounded in observed training behavior while preserving the executable structure of the previous design.

\begin{quantprompt}{DGR Prompt --- Diagnostic-Guided Refinement}
    \scriptsize
    \linespread{0.90}\selectfont
    \color{quanttext}

    \promptlabel{Role:}
    \promptcontent{Refine the state enhancement, intrinsic reward, and risk-rule configuration using PPO training diagnostics. Do not output actions or portfolio weights.}\\[-1pt]

    \promptlabel{Inputs:}
    \promptcontent{Previous}
    \promptcode{revise\_state(s)}
    \promptcontent{,}
    \promptcode{intrinsic\_reward(updated\_s)}
    \promptcontent{, selected risk rules, market context, and rollout diagnostics.}\\[-1pt]

    \promptlabel{Diagnostics:}
    \promptcontent{Feature channels are summarized by predictive correlation, critic reliance, and regime-specific behavior. The intrinsic reward is summarized by scale, stability, and correlation with policy performance.}\\[-1pt]

    \promptlabel{Revision Rules:}
    \promptcontent{Retain predictive features used by the critic; reduce weak features that the critic overuses; make underused but predictive features more explicit; remove or invert negatively aligned features. If volatile-regime performance is weak, add risk-sensitive features or strengthen defensive reward terms.}\\[-1pt]

    \promptlabel{Risk Adjustment:}
    \promptcontent{Increase drawdown or defensive rules when the policy has high drawdown. Reduce redundant risk penalties when the policy is overly conservative.}\\[-1pt]

    \promptlabel{Output:}
    \promptcontent{Return next-round code for}
    \promptcode{revise\_state(s)}
    \promptcontent{and}
    \promptcode{intrinsic\_reward(updated\_s)}
    \promptcontent{, selected risk rules, \(\lambda\), and a concise rationale.}
\end{quantprompt}

The diagnostic report injected into DGR follows the compact format below. 
The values are illustrative; actual values are computed from the corresponding PPO-trained candidate.

\begin{minipage}{\linewidth}
\begin{lstlisting}[style=GIFTplain]
========== Candidate i ==========
Training summary:
  Sharpe = ...
  Return = ...%
  MaxDD  = ...

Feature diagnostics:
  s[120]: predictive corr = +0.0742, critic reliance = high
  s[121]: predictive corr = -0.0413, critic reliance = medium
  s[122]: predictive corr = +0.0091, critic reliance = high

Regime diagnostics:
  volatile:   s[121] negative, risk sensitivity weak
  trending_up: s[120] positive

Intrinsic-reward diagnostics:
  mean = ...
  scale = ...
  corr with policy performance = ...

Suggested revision:
  - retain or extend s[120];
  - remove or invert s[121];
  - reduce s[122] if weakly predictive;
  - add a risk-sensitive feature if volatile-regime performance is weak.
\end{lstlisting}
\end{minipage}

\section{Experimental Setup and Portfolio Construction}
\label{app:exp_setup}

This section provides additional details on the market data, rolling-window protocol, portfolio construction, baseline setting, and candidate-selection procedure used in our experiments. 
The purpose is to clarify the evaluation design behind GIFT while keeping the Appendix aligned with the main paper: GIFT uses the LLM to design a state-reward interface for PPO, rather than to output trading actions or portfolio weights.

\subsection{Market Data and Trading Environment}
\label{app:data_environment}

We use daily market data from U.S.-listed S\&P 500 equities, covering January 2000 to June 2024.
For each asset, the raw fields include open, high, low, close, volume, and adjusted close prices.
Daily returns and portfolio net values are computed from adjusted prices when available, while OHLCV observations form the raw market input to the PPO agent.
All feature computations are causal: at each decision time, price-derived inputs only use information available up to that day.

Each portfolio panel contains five risky assets and one cash position.
At each trading day, the policy observes a 20-day OHLCV-related history and current portfolio context, and outputs a long-only allocation over the five stocks and cash:
\begin{equation}
    w_{t,i}\ge 0,\qquad
    w_{t,\mathrm{cash}}\ge 0,\qquad
    \sum_{i=1}^{N} w_{t,i}+w_{t,\mathrm{cash}}=1 ,
\end{equation}
where \(i=1,\ldots,N\) denotes the risky assets, \(N=5\), and the cash position is stored as the final action dimension.
We apply a proportional transaction cost of \(0.1\%\) per unit turnover.
All methods use the same preprocessing, market-calendar alignment, action constraint, transaction-cost setting, and evaluation windows.

\begin{table}[t]
\centering
\scriptsize
\setlength{\tabcolsep}{4pt}
\renewcommand{\arraystretch}{1.10}
\caption{
Summary of the market data and trading environment.
}
\label{tab:dataset_summary}
\resizebox{\columnwidth}{!}{
\begin{tabular}{ll}
\toprule
\textbf{Item} & \textbf{Description} \\
\midrule
Market & U.S. S\&P 500 equities \\
Data range & January 2000--June 2024 \\
Frequency & Daily trading data \\
Raw fields & Open, high, low, close, volume, adjusted close \\
Return computation & Daily returns from adjusted prices when available \\
Portfolio unit & Five stocks plus cash \\
Action space & Long-only simplex allocation \\
Lookback window & 20 trading days \\
Transaction cost & 0.1\% per unit turnover \\
Evaluation protocol & Rolling train/evaluation split \\
\bottomrule
\end{tabular}
}
\end{table}

\subsection{Rolling-window Evaluation Protocol}
\label{app:rolling_windows}

We evaluate all methods under six rolling test windows, denoted as W1--W6.
The windows cover post-pandemic recovery, late bull-market conditions, rate-hike and geopolitical shocks, high-rate consolidation, banking stress with an AI-led rebound, and the second-half 2023 recovery.
These regimes are used only to contextualize results; regime labels are not used as model inputs.

For each window, GIFT performs LLM-guided interface search only on historical data before the reported evaluation segment.
The selected state-enhancement and reward-shaping interface is then frozen.
During evaluation, the LLM is not queried, the prompt is not updated, the feedback memory is not modified, and the selected interface is not changed.
A newly initialized PPO agent adapts on the first half of the test window using the frozen interface and is evaluated on the second half.
The Pure PPO baseline follows the same chronological split and PPO training protocol, but keeps the original raw state and portfolio-return reward.

\subsection{Portfolio Panel Construction}
\label{app:portfolio_panels}

The evaluation unit is a portfolio panel.
As Table~\ref{tab:portfolio_panels_and_stock_characteristics} shows, panel contains five representative S\&P 500 stocks and one cash position, forming a long-only simplex-constrained asset pool.
Single-sector panels test whether the generated state-reward interface remains useful under different sector characteristics, while mixed panels introduce broader cross-sector heterogeneity.
These panels are controlled evaluation settings and are not intended to represent the entire U.S. equity market.

\begin{table*}[htbp]
\centering
\footnotesize
\setlength{\tabcolsep}{0.8pt}
\renewcommand{\arraystretch}{0.68}
\setlength{\extrarowheight}{0.8pt}
\caption{
Portfolio panels and stock-level characteristics used in the rolling-window evaluation.
Each panel contains five risky assets and one cash position under the same long-only simplex constraint.
}
\label{tab:portfolio_panels_and_stock_characteristics}

\begin{tabularx}{\textwidth}{
    >{\raggedright\arraybackslash}p{0.14\textwidth}
    >{\raggedright\arraybackslash}p{0.19\textwidth}
    >{\raggedright\arraybackslash}p{0.29\textwidth}
    >{\raggedright\arraybackslash}X
}
\toprule
\textbf{Panel / Group} 
& \textbf{Ticker or Constituents} 
& \textbf{Broad Exposure} 
& \textbf{Role in Evaluation} \\
\midrule

\multicolumn{4}{l}{\textit{Single-sector portfolio panels}} \\
\midrule
Technology 
& AAPL, MSFT, NVDA, GOOGL, META
& Large-cap technology and platform firms
& Tests growth-oriented, high-correlation technology exposure. \\

Healthcare 
& JNJ, LLY, UNH, MRK, PFE
& Pharmaceuticals, healthcare services, and medical products
& Tests a more defensive sector with distinct risk drivers. \\

Energy 
& XOM, CVX, COP, SLB, EOG
& Integrated energy, upstream production, and oilfield services
& Tests commodity-sensitive exposure to energy cycles. \\

Industrials
& BA, CAT, UNP, LMT, WM
& Aerospace, machinery, rail transportation, defense, and industrial services
& Tests cyclical industrial exposure linked to aerospace, infrastructure demand, logistics, defense cycles, and industrial services. \\

\addlinespace[3pt]
\midrule
\multicolumn{4}{l}{\textit{Mixed portfolio panels}} \\
\midrule
Light Mix 
& TSLA, NFLX, AMZN, MSFT, JNJ
& Technology and consumer-growth assets with one healthcare anchor
& Tests moderately mixed exposure with a defensive component. \\

Heavy Mix 
& TSLA, NVDA, XOM, CAT, JNJ
& Technology, energy, industrials, and healthcare
& Tests stronger cross-sector heterogeneity. \\

\addlinespace[3pt]
\midrule
\multicolumn{4}{l}{\textit{Technology and platform stocks}} \\
\midrule
Technology 
& AAPL 
& Consumer electronics and platform services
& Mature technology anchor. \\

Technology 
& MSFT 
& Enterprise software, cloud, and AI infrastructure
& Stable technology anchor; used in Technology and Light Mix. \\

Technology 
& NVDA 
& Semiconductors, GPUs, and AI infrastructure
& High-growth semiconductor exposure; used in Technology and Heavy Mix. \\

Technology 
& GOOGL 
& Search, advertising, cloud, and internet platforms
& Platform stock exposed to advertising and cloud growth. \\

Technology 
& META 
& Social platforms and digital advertising
& Consumer-internet stock with advertising-cycle exposure. \\

\addlinespace[3pt]
\midrule
\multicolumn{4}{l}{\textit{Healthcare stocks}} \\
\midrule
Healthcare 
& JNJ 
& Diversified healthcare and medical products
& Defensive healthcare anchor; used in Healthcare, Light Mix, and Heavy Mix. \\

Healthcare 
& LLY 
& Pharmaceuticals and specialty drugs
& Growth-oriented healthcare exposure. \\

Healthcare 
& UNH 
& Managed care and healthcare services
& Healthcare-services stock distinct from drug manufacturers. \\

Healthcare 
& MRK 
& Pharmaceuticals, oncology, and vaccines
& Mature pharmaceutical exposure. \\

Healthcare 
& PFE 
& Pharmaceuticals and biopharmaceutical products
& Drug-cycle and pipeline-sensitive healthcare exposure. \\

\addlinespace[3pt]
\midrule
\multicolumn{4}{l}{\textit{Energy stocks}} \\
\midrule
Energy 
& XOM 
& Integrated oil and gas
& Large-cap energy anchor; used in Energy and Heavy Mix. \\

Energy 
& CVX 
& Integrated oil and gas
& Broad oil-and-gas price exposure. \\

Energy 
& COP 
& Oil and gas exploration and production
& Upstream producer with direct commodity exposure. \\

Energy 
& SLB 
& Oilfield services and energy equipment
& Energy-services exposure distinct from integrated producers. \\

Energy 
& EOG 
& Oil and gas exploration and production
& Upstream producer enriching commodity-linked risk. \\

\addlinespace[3pt]
\midrule
\multicolumn{4}{l}{\textit{Industrials stocks}} \\
\midrule
Industrials 
& BA
& Aerospace and aircraft manufacturing
& Adds aerospace and aviation-cycle exposure. \\

Industrials 
& CAT 
& Machinery, construction, mining, and infrastructure demand
& Cyclical industrial anchor; used in Industrials and Heavy Mix. \\

Industrials 
& UNP 
& Rail transportation and freight logistics
& Adds transportation and logistics-cycle exposure. \\

Industrials 
& LMT 
& Defense, aerospace, and government-linked industrial demand
& Adds defense-oriented industrial exposure with distinct risk drivers. \\

Industrials 
& WM
& Waste management and environmental services
& Adds industrial-services exposure with relatively defensive demand characteristics. \\

\addlinespace[3pt]
\midrule
\multicolumn{4}{l}{\textit{Additional mixed-panel stocks}} \\
\midrule
Consumer / Growth 
& TSLA 
& Electric vehicles and growth-oriented consumer technology
& High-volatility growth anchor in Light Mix and Heavy Mix. \\

Consumer / Media 
& NFLX 
& Streaming media and subscription entertainment
& Adds media and consumer-growth exposure in Light Mix. \\

Consumer / Cloud 
& AMZN 
& E-commerce, cloud, and logistics
& Hybrid consumer-technology exposure in Light Mix. \\

\bottomrule
\end{tabularx}
\end{table*}

\subsection{Baseline and Evaluation Metrics}
\label{app:baseline_metrics}

We compare GIFT with a Pure PPO baseline under the same rolling-window protocol.
The baseline uses the same asset universe, transaction cost, adaptation period, evaluation period, PPO backbone, and portfolio constraints.
Its only difference from GIFT is that it does not use LLM-generated state enhancement, reward shaping, or diagnostic-guided interface refinement.
This comparison isolates the contribution of the generated state-reward interface.

We report five standard portfolio metrics.
Cumulative return measures total portfolio profit over the evaluation period.
Sharpe ratio measures risk-adjusted return using total return volatility.
Sortino ratio focuses on downside volatility.
Maximum drawdown (MDD) measures the largest peak-to-trough loss and reflects downside risk.
Calmar ratio compares return against maximum drawdown.
Cumulative return, Sharpe, Sortino, and Calmar are better when higher, while MDD is better when lower.

For behavioral analysis, we also report effective diversification and turnover.
Effective diversification is computed as
\begin{equation}
    N_{\mathrm{eff}} = \frac{1}{\sum_i w_i^2},
\end{equation}
where larger values indicate less concentrated allocations.
Turnover is computed from the average \(\ell_1\) change in portfolio weights across adjacent trading days and is used to diagnose trading-cost sensitivity.

\subsection{Rollout Scoring and Candidate Selection}
\label{app:rollout_scoring}

During interface search, GIFT ranks candidate state-reward interfaces by PPO rollout performance on the historical search window.
For candidate \(k\) in refinement round \(m\), the generated state interface \(F_{m,k}\) and reward interface \(R_{m,k}\) are attached to the portfolio environment, and PPO is trained on the search data.
Let \(\mathcal{R}_{m,k}^{\mathrm{train}}=\{r_t\}_{t=1}^{T_{m,k}}\) denote the portfolio-return sequence collected during this rollout.
The rollout score is the annualized Sharpe ratio:
\begin{equation}
S^{\mathrm{ro}}_{m,k}
=
\frac{252\,\overline{r}_{m,k}-r_f}
{\sqrt{252}\,\sigma(r_{m,k})},
\end{equation}
where \(r_f=0\) in our implementation.
If the return sequence has fewer than two observations or zero variance, the score is set to zero.

Within each refinement round, GIFT selects the candidate with the highest rollout score:
\begin{equation}
k_m^\star
=
\arg\max_k S^{\mathrm{ro}}_{m,k}.
\end{equation}
The selected candidate supplies both the performance record and the diagnostics used for the next refinement step, including return, drawdown, risk-adjusted performance, feature informativeness, and reward-stability summaries.
Across refinement rounds, GIFT tracks the best candidate observed so far:
\begin{equation}
(F^\star,R^\star)
=
\arg\max_{m,k} S^{\mathrm{ro}}_{m,k}.
\end{equation}

Before any generated interface is used for PPO training, it must pass basic validity checks.
The state-enhancement function must preserve the raw market input, append only finite numerical features, and return a valid one-dimensional state representation.
The intrinsic reward must return a finite scalar within a bounded range, and reward-rule coefficients are constrained to predefined safe ranges.
These checks are used only to ensure executable and numerically stable candidate interfaces; they are not used to tune the test-window policy.

During test-window evaluation, GIFT does not use rollout scores to reselect or modify the interface.
The stored best interface is attached to a newly initialized PPO agent, which adapts on the first half of the test window and is evaluated on the second half.
No LLM query, prompt update, feedback-memory write, or interface modification is allowed during this stage.

\section{Additional Comparison with Traditional Trading Baselines}
\label{app:traditional_trading_baselines}

\paragraph{Representative trading baselines.}
Although the main experiments focus on whether GIFT improves PPO through state-reward interface design, we also compare it with several traditional trading baselines as an external reference:

\begin{itemize}[leftmargin=1.3em, itemsep=1pt, topsep=2pt, parsep=0pt]
    \item \textbf{SMA/WMA:} moving-average crossover rules for trend-following signals ~\cite{mcconnell2008equity}.
    \item \textbf{ATR/Bollinger Bands:} volatility-band strategies for breakout or deviation-based trading~\cite{bollinger2002bollinger}.
    \item \textbf{Turn-of-the-Month:} a calendar-effect strategy based on return concentration around month boundaries~\cite{mcconnell2008equity}.
    \item \textbf{XGBoost:} a gradient-boosted tree forecaster that converts predicted market direction into trading signals~\cite{chen2016xgboost}.
\end{itemize}

\paragraph{\bfseries\itshape Evaluation protocol.}
This comparison provides an external reference beyond the controlled pure-PPO comparison in the main text. 
We compare GIFT with six baselines: SMA, WMA, ATR, Bollinger Bands, Turn-of-the-Month, and XGBoost.
The evaluation covers six portfolio panels: four sector panels (Technology, Healthcare, Energy, and Industrials) and two mixed panels (Light Mix and Heavy Mix). 
Each panel uses its own five-stock universe.

GIFT follows the same offline interface-design protocol as in the main experiments. 
The selected state-enhancement and reward-shaping code is fixed before evaluation, with no LLM query, prompt update, feedback-memory write, or interface modification during evaluation.

Table~\ref{tab:gift_vs_trading_baselines} reports six-window averages of cumulative return, Sharpe, Sortino, maximum drawdown (MDD), and Calmar. 
Higher values are better except for MDD.
For each baseline row, a checkmark indicates that GIFT performs better on that metric, while a cross indicates that the baseline performs better. 
The Wins column reports the number of metrics won by GIFT; majority-metric wins, i.e., at least 3 out of 5 metrics, are highlighted in bold.

\begin{table*}[htbp]
\centering
\caption{
Six-window average comparison between GIFT and traditional trading baselines. 
Green checkmarks denote GIFT wins, red crosses denote baseline wins, and bold Wins indicate majority-metric wins ($\geq 3/5$).
Best values are bolded and second-best values are underlined within each panel.
}
\label{tab:gift_vs_trading_baselines}

\tiny
\setlength{\tabcolsep}{1.4pt}
\renewcommand{\arraystretch}{0.86}
\setlength{\aboverulesep}{0.25pt}
\setlength{\belowrulesep}{0.25pt}
\setlength{\cmidrulesep}{0.25pt}
\resizebox{0.85\textwidth}{!}{
\begin{tabular}{l *{5}{r@{\hspace{0.08em}}c} c}

\toprule
\textbf{Method}
& \multicolumn{2}{c}{\textbf{Ret.\ (\%)$\uparrow$}}
& \multicolumn{2}{c}{\textbf{Sharpe $\uparrow$}}
& \multicolumn{2}{c}{\textbf{Sortino $\uparrow$}}
& \multicolumn{2}{c}{\textbf{MDD (\%)$\downarrow$}}
& \multicolumn{2}{c}{\textbf{Calmar $\uparrow$}}
& \textbf{Wins} \\
\midrule

\multicolumn{12}{l}{\textit{(a) Technology (AAPL, MSFT, NVDA, GOOGL, META)}} \\
\midrule
SMA
& 0.85 & \gcheck
& 0.09 & \gcheck
& 0.22 & \gcheck
& 10.45 & \rcross
& 0.32 & \gcheck
& \textbf{4/5} \\
WMA
& -0.18 & \gcheck
& -0.04 & \gcheck
& 0.01 & \gcheck
& 9.91 & \rcross
& 0.17 & \gcheck
& \textbf{4/5} \\
ATR
& 3.01 & \gcheck
& 0.51 & \gcheck
& 0.39 & \gcheck
& 13.86 & \gcheck
& 1.03 & \gcheck
& \textbf{5/5} \\
Bollinger
& 2.52 & \gcheck
& 0.47 & \gcheck
& 0.38 & \gcheck
& 10.95 & \rcross
& 0.78 & \gcheck
& \textbf{4/5} \\
TurnMonth
& \best{12.46} & \rcross
& 0.70 & \gcheck
& 0.59 & \gcheck
& \second{9.86} & \rcross
& 1.38 & \gcheck
& \textbf{3/5} \\
XGBoost
& 6.86 & \gcheck
& \second{0.86} & \gcheck
& \second{0.95} & \gcheck
& \best{8.12} & \rcross
& \second{2.08} & \gcheck
& \textbf{4/5} \\
\textbf{GIFT (Ours)}
& \second{9.17} & 
& \best{1.14} &
& \best{1.48} &
& 13.74 &
& \best{4.09} &
& --- \\

\midrule
\multicolumn{12}{l}{\textit{(b) Healthcare (JNJ, LLY, UNH, MRK, PFE)}} \\
\midrule
SMA
& 5.53 & \gcheck
& 0.66 & \gcheck
& 0.91 & \gcheck
& 6.59 & \gcheck
& 1.08 & \gcheck
& \textbf{5/5} \\
WMA
& \second{8.94} & \gcheck
& \second{1.02} & \gcheck
& \second{1.38} & \gcheck
& 6.16 & \rcross
& 1.59 & \gcheck
& \textbf{4/5} \\
ATR
& 3.11 & \gcheck
& 0.35 & \gcheck
& 0.47 & \gcheck
& 6.01 & \rcross
& 0.45 & \gcheck
& \textbf{4/5} \\
Bollinger
& 6.24 & \gcheck
& 0.80 & \gcheck
& 1.22 & \gcheck
& \second{5.71} & \rcross
& \second{1.74} & \gcheck
& \textbf{4/5} \\
TurnMonth
& 5.89 & \gcheck
& 0.61 & \gcheck
& 0.56 & \gcheck
& 6.61 & \gcheck
& 0.75 & \gcheck
& \textbf{5/5} \\
XGBoost
& 6.06 & \gcheck
& 0.82 & \gcheck
& 0.92 & \gcheck
& \best{4.03} & \rcross
& 1.31 & \gcheck
& \textbf{4/5} \\
\textbf{GIFT (Ours)}
& \best{9.35} &
& \best{1.63} &
& \best{1.84} &
& 6.36 &
& \best{4.03} &
& --- \\

\midrule
\multicolumn{12}{l}{\textit{(c) Energy (XOM, CVX, COP, SLB, EOG)}} \\
\midrule
SMA
& -5.55 & \gcheck
& -0.30 & \gcheck
& -0.31 & \gcheck
& 18.76 & \gcheck
& -0.15 & \gcheck
& \textbf{5/5} \\
WMA
& 1.08 & \gcheck
& 0.09 & \gcheck
& 0.14 & \gcheck
& 13.95 & \gcheck
& 0.12 & \gcheck
& \textbf{5/5} \\
ATR
& \best{18.38} & \rcross
& \second{1.04} & \gcheck
& \second{1.08} & \gcheck
& \second{9.98} & \rcross
& 1.52 & \gcheck
& \textbf{3/5} \\
Bollinger
& 8.01 & \gcheck
& 0.52 & \gcheck
& 0.52 & \gcheck
& 10.43 & \rcross
& 0.64 & \gcheck
& \textbf{4/5} \\
TurnMonth
& \second{17.93} & \rcross
& 0.86 & \gcheck
& 0.72 & \gcheck
& 10.14 & \rcross
& \second{1.73} & \gcheck
& \textbf{3/5} \\
XGBoost
& 8.83 & \gcheck
& 0.93 & \gcheck
& 0.83 & \gcheck
& \best{6.53} & \rcross
& 1.46 & \gcheck
& \textbf{4/5} \\
\textbf{GIFT (Ours)}
& 12.03 &
& \best{1.09} &
& \best{1.21} &
& 13.24 &
& \best{3.01} &
& --- \\

\midrule
\multicolumn{12}{l}{\textit{(d) Industrials (BA, CAT, UNP, LMT, WM)}} \\
\midrule
SMA
& 0.55 & \gcheck
& 0.06 & \gcheck
& 0.14 & \gcheck
& 9.92 & \gcheck
& 0.15 & \gcheck
& \textbf{5/5} \\
WMA
& 4.11 & \gcheck
& 0.42 & \gcheck
& 0.51 & \gcheck
& 8.02 & \rcross
& 0.53 & \gcheck
& \textbf{4/5} \\
ATR
& 6.76 & \gcheck
& 0.66 & \gcheck
& 0.74 & \gcheck
& 7.76 & \rcross
& 0.76 & \gcheck
& \textbf{4/5} \\
Bollinger
& 3.43 & \gcheck
& 0.32 & \gcheck
& 0.41 & \gcheck
& 8.82 & \gcheck
& 0.50 & \gcheck
& \textbf{5/5} \\
TurnMonth
& \best{10.94} & \rcross
& 0.99 & \gcheck
& 0.85 & \gcheck
& \second{6.63} & \rcross
& 1.46 & \gcheck
& \textbf{3/5} \\
XGBoost
& \second{8.01} & \rcross
& \second{1.24} & \gcheck
& \second{1.33} & \gcheck
& \best{3.62} & \rcross
& \second{1.73} & \gcheck
& \textbf{3/5} \\
\textbf{GIFT (Ours)}
& 7.24 &
& \best{1.44} &
& \best{1.54} &
& 8.54 &
& \best{2.83} &
& --- \\

\midrule
\multicolumn{12}{l}{\textit{(e) Light Mix (TSLA, NFLX, AMZN, MSFT, JNJ)}} \\
\midrule
SMA
& 3.96 & \rcross
& 0.22 & \gcheck
& 0.27 & \gcheck
& 11.27 & \rcross
& 0.30 & \gcheck
& \textbf{3/5} \\
WMA
& 5.96 & \rcross
& 0.40 & \gcheck
& 0.51 & \gcheck
& \second{9.51} & \rcross
& 0.63 & \gcheck
& \textbf{3/5} \\
ATR
& 0.58 & \gcheck
& 0.34 & \gcheck
& 0.33 & \gcheck
& 13.33 & \gcheck
& 0.71 & \gcheck
& \textbf{5/5} \\
Bollinger
& 2.81 & \gcheck
& 0.43 & \gcheck
& 0.38 & \gcheck
& 14.11 & \gcheck
& 0.56 & \gcheck
& \textbf{5/5} \\
TurnMonth
& \best{8.48} & \rcross
& 0.55 & \gcheck
& 0.47 & \gcheck
& \best{8.54} & \rcross
& 1.07 & \gcheck
& \textbf{3/5} \\
XGBoost
& \second{6.55} & \rcross
& \second{0.61} & \gcheck
& \second{0.67} & \gcheck
& 9.63 & \rcross
& \second{1.12} & \gcheck
& \textbf{3/5} \\
\textbf{GIFT (Ours)}
& 3.00 &
& \best{0.73} &
& \best{0.78} &
& 11.78 &
& \best{1.38} &
& --- \\

\midrule
\multicolumn{12}{l}{\textit{(f) Heavy Mix (TSLA, NVDA, XOM, CAT, JNJ)}} \\
\midrule
SMA
& 7.94 & \rcross
& 0.57 & \gcheck
& 0.78 & \gcheck
& 8.35 & \rcross
& 1.12 & \gcheck
& \textbf{3/5} \\
WMA
& \second{12.38} & \rcross
& 0.77 & \gcheck
& \second{1.01} & \gcheck
& 8.61 & \rcross
& \second{1.65} & \gcheck
& \textbf{3/5} \\
ATR
& 5.32 & \gcheck
& 0.52 & \gcheck
& 0.55 & \gcheck
& 10.28 & \gcheck
& 0.76 & \gcheck
& \textbf{5/5} \\
Bollinger
& 8.32 & \rcross
& 0.73 & \gcheck
& 0.89 & \gcheck
& 9.39 & \rcross
& 0.96 & \gcheck
& \textbf{3/5} \\
TurnMonth
& \best{13.04} & \rcross
& \second{0.91} & \gcheck
& 0.70 & \gcheck
& \second{7.44} & \rcross
& 1.64 & \gcheck
& \textbf{3/5} \\
XGBoost
& 8.19 & \rcross
& 0.87 & \gcheck
& 0.91 & \gcheck
& \best{6.57} & \rcross
& 1.41 & \gcheck
& \textbf{3/5} \\
\textbf{GIFT (Ours)}
& 7.26 &
& \best{1.08} &
& \best{1.21} &
& 10.20 &
& \best{2.96} &
& --- \\

\bottomrule
\end{tabular}
}
\end{table*}

As shown in Table~\ref{tab:gift_vs_trading_baselines}, GIFT achieves majority-metric wins in all head-to-head comparisons under this summary protocol. 
The advantage is more consistent on risk-adjusted metrics, especially Sharpe, Sortino, and Calmar, than on raw cumulative return or MDD. 
This suggests that the GIFT-designed state-reward interface mainly helps PPO improve return-risk trade-offs, rather than simply increasing cumulative return.

The mixed panels show a similar pattern. 
Several rule-based or predictor-based baselines obtain higher cumulative return or lower MDD in some cases, but GIFT often retains stronger risk-adjusted performance. 
This is consistent with the role of GIFT as an interface-design method for PPO: it does not encode a fixed trading rule, but provides a structured state-reward interface from which the PPO policy learns.

\section{Detailed Results of state-reward Interface Variants}
\label{app:interface_variant_details}

This section provides the full per-metric results underlying
Table~\ref{tab:variant_stability}.
All variants are evaluated with the same PPO backbone and chronological
rolling-window protocol as Pure PPO.
The comparison isolates how the state-reward interface is constructed:
fixed factor-state enhancement, fixed risk-rule reward shaping, a fixed
state-reward interface, free-form LLM interface generation, and GIFT.
For each rolling window, a metric is counted as a win if the method improves
over Pure PPO on that metric.
Sharpe, Sortino, Calmar, and Return are better when higher, whereas MDD is
better when lower; ties are not counted as wins.

\begin{table*}[p]
\centering
\tiny
\setlength{\tabcolsep}{3.6pt}
\renewcommand{\arraystretch}{1.06}
\caption{
Detailed per-metric results of state-reward interface variants across six rolling windows.
All variants use the same PPO backbone and rolling-window protocol.
Bold values indicate improvements over Pure PPO on the corresponding metric.
}
\label{tab:interface_variant_details}
\resizebox{\textwidth}{!}{
\begin{tabular}{llrrrrrc}
\toprule
\textbf{Window}
& \textbf{Method}
& \textbf{Sharpe$\uparrow$}
& \textbf{Sortino$\uparrow$}
& \textbf{MDD(\%)$\downarrow$}
& \textbf{Calmar$\uparrow$}
& \textbf{Return(\%)$\uparrow$}
& \textbf{Wins} \\
\midrule

\multirow{6}{*}{W1}
& Pure PPO
& -0.143 & -0.140 & 11.93 & -0.221 & -1.78 & -- \\
& Fixed factor-state enhancement
& \textbf{1.752} & \textbf{1.606} & 11.93 & \textbf{14.69} & \textbf{42.90} & 4/5 \\
& Fixed risk-rule reward shaping
& \textbf{1.545} & \textbf{1.605} & \textbf{11.81} & \textbf{13.08} & \textbf{34.59} & 5/5 \\
& Fixed state-reward interface
& \textbf{2.057} & \textbf{1.931} & \textbf{10.22} & \textbf{20.13} & \textbf{41.22} & 5/5 \\
& Free-form LLM interface
& \textbf{0.631} & \textbf{0.616} & \textbf{10.13} & \textbf{1.153} & \textbf{5.07} & 5/5 \\
& \textbf{GIFT}
& \textbf{0.167} & \textbf{0.170} & \textbf{11.67} & \textbf{0.258} & \textbf{0.57} & \textbf{5/5} \\

\midrule

\multirow{6}{*}{W2}
& Pure PPO
& 2.083 & 2.116 & 7.65 & 3.753 & 12.13 & -- \\
& Fixed factor-state enhancement
& 1.225 & 1.110 & 10.16 & \textbf{12.06} & \textbf{18.41} & 2/5 \\
& Fixed risk-rule reward shaping
& 1.223 & 1.054 & 12.26 & \textbf{9.97} & \textbf{19.35} & 2/5 \\
& Fixed state-reward interface
& 1.164 & \textbf{2.294} & 10.71 & \textbf{10.87} & \textbf{15.94} & 3/5 \\
& Free-form LLM interface
& 1.758 & 1.638 & 7.69 & 2.918 & 9.61 & 0/5 \\
& \textbf{GIFT}
& \textbf{2.183} & \textbf{2.192} & \textbf{5.53} & \textbf{4.862} & 11.37 & \textbf{4/5} \\

\midrule

\multirow{6}{*}{W3}
& Pure PPO
& -2.048 & -1.720 & 35.98 & -2.406 & -32.70 & -- \\
& Fixed factor-state enhancement
& \textbf{2.392} & \textbf{2.459} & \textbf{8.38} & \textbf{28.55} & \textbf{15.82} & 5/5 \\
& Fixed risk-rule reward shaping
& \textbf{2.266} & \textbf{2.330} & \textbf{7.40} & \textbf{30.63} & \textbf{13.56} & 5/5 \\
& Fixed state-reward interface
& \textbf{2.121} & \textbf{2.432} & \textbf{8.25} & \textbf{25.71} & \textbf{12.83} & 5/5 \\
& Free-form LLM interface
& \textbf{-1.698} & \textbf{-1.559} & \textbf{28.10} & \textbf{-1.783} & \textbf{-23.70} & 5/5 \\
& \textbf{GIFT}
& \textbf{-1.497} & \textbf{-1.335} & \textbf{23.43} & \textbf{-1.775} & \textbf{-17.13} & \textbf{5/5} \\

\midrule

\multirow{6}{*}{W4}
& Pure PPO
& -2.304 & -2.419 & 26.35 & -2.322 & -23.46 & -- \\
& Fixed factor-state enhancement
& \textbf{-1.913} & \textbf{-1.534} & 30.99 & -6.17 & -26.68 & 2/5 \\
& Fixed risk-rule reward shaping
& \textbf{-1.700} & \textbf{-1.372} & 29.18 & -5.83 & \textbf{-22.97} & 3/5 \\
& Fixed state-reward interface
& \textbf{-1.272} & \textbf{-1.011} & \textbf{22.84} & -5.57 & \textbf{-15.81} & 4/5 \\
& Free-form LLM interface
& \textbf{-2.052} & \textbf{-2.171} & 28.67 & \textbf{-2.042} & \textbf{-23.18} & 4/5 \\
& \textbf{GIFT}
& \textbf{-0.657} & \textbf{-0.700} & \textbf{13.46} & \textbf{-1.383} & \textbf{-8.91} & \textbf{5/5} \\

\midrule

\multirow{6}{*}{W5}
& Pure PPO
& 1.986 & 2.321 & 11.33 & 2.851 & 13.62 & -- \\
& Fixed factor-state enhancement
& -0.849 & -1.041 & 20.00 & -4.25 & -13.33 & 0/5 \\
& Fixed risk-rule reward shaping
& -1.122 & -1.511 & 21.72 & -5.17 & -15.19 & 0/5 \\
& Fixed state-reward interface
& -0.920 & -1.104 & 17.28 & -5.32 & -11.19 & 0/5 \\
& Free-form LLM interface
& \textbf{2.657} & \textbf{2.885} & \textbf{8.69} & \textbf{5.097} & \textbf{23.47} & 5/5 \\
& \textbf{GIFT}
& \textbf{2.889} & \textbf{3.196} & \textbf{9.00} & \textbf{4.860} & \textbf{19.21} & \textbf{5/5} \\

\midrule

\multirow{6}{*}{W6}
& Pure PPO
& 0.785 & 0.771 & 12.28 & 1.158 & 5.28 & -- \\
& Fixed factor-state enhancement
& \textbf{2.600} & \textbf{2.749} & \textbf{10.90} & \textbf{23.86} & \textbf{22.54} & 5/5 \\
& Fixed risk-rule reward shaping
& \textbf{2.702} & \textbf{2.961} & \textbf{11.02} & \textbf{24.51} & \textbf{24.49} & 5/5 \\
& Fixed state-reward interface
& \textbf{2.856} & \textbf{3.125} & \textbf{9.32} & \textbf{30.64} & \textbf{22.60} & 5/5 \\
& Free-form LLM interface
& 0.148 & 0.139 & 12.74 & 0.174 & 0.46 & 0/5 \\
& \textbf{GIFT}
& \textbf{0.996} & \textbf{0.961} & \textbf{10.67} & \textbf{1.354} & \textbf{5.63} & \textbf{5/5} \\

\bottomrule
\end{tabular}
}
\end{table*}

Table~\ref{tab:interface_variant_details} shows that each interface strategy has
regime-dependent behavior.
Fixed factor-state enhancement and fixed reward shaping can improve Pure PPO in
some windows, but they become unstable when the market pattern changes; notably,
all fixed variants lose on all five metrics in W5.
The free-form LLM interface is also inconsistent, achieving full wins in some
windows but failing to improve Pure PPO in W2 and W6.
In contrast, GIFT obtains at least four wins in every window and 29 out of 30
metric-level wins overall.
These results support the RQ2 finding that the main benefit comes from combining
financially constrained interface generation with diagnostic-guided refinement,
rather than relying on fixed designs or unconstrained LLM generation alone.

\section{Ablation Details}
\label{app:ablation_details}

Table~\ref{tab:ablation_main} reports the average performance change after removing each component from GIFT.
This section provides the per-window results used to compute these averaged changes.
All variants use the same PPO backbone, rolling-window protocol, and evaluation metrics.
The full model keeps Factor-guided State Enhancement (FSE), Risk-rule-guided Reward Shaping (RRS), and Diagnostic-guided Refinement (DGR).

\paragraph{\bfseries\itshape Ablation settings.}
\textit{w/o FSE} removes the LLM-generated state-enhancement function, so PPO uses the original market state without appended factor channels.
\textit{w/o RRS} removes both the LLM-generated intrinsic reward and the risk-rule rewards, leaving PPO without auxiliary reward shaping.
\textit{w/o DGR} disables iterative diagnostic refinement and uses a one-round interface generation process.
\textit{w/o IC/SHAP} keeps the refinement loop but removes factor-level IC and critic-attribution diagnostics from the feedback signal.

\begin{table}[htbp]
\centering
\tiny
\setlength{\tabcolsep}{2.4pt}
\renewcommand{\arraystretch}{1.04}
\caption{
Per-window ablation results across all evaluation metrics.
Return and MDD are reported in percentages.
}
\label{tab:ablation_details_all}
\resizebox{\columnwidth}{!}{
\begin{tabular}{llrrrrrr}
\toprule
\textbf{Metric}
& \textbf{Method}
& \textbf{W1}
& \textbf{W2}
& \textbf{W3}
& \textbf{W4}
& \textbf{W5}
& \textbf{W6} \\
\midrule

\multirow{5}{*}{Sharpe$\uparrow$}
& Full GIFT
& 0.447 & 2.107 & -1.247 & -0.723 & 2.712 & 1.114 \\
& w/o FSE
& 0.614 & 2.028 & -1.347 & -1.440 & 2.515 & 1.267 \\
& w/o RRS
& -0.312 & 2.191 & -1.127 & -1.691 & 2.853 & 1.720 \\
& w/o DGR
& 0.451 & 1.089 & -1.255 & -2.013 & 2.707 & 1.071 \\
& w/o IC/SHAP
& -0.026 & 2.353 & -1.611 & -1.487 & 2.414 & 1.224 \\
\midrule

\multirow{5}{*}{Sortino$\uparrow$}
& Full GIFT
& 0.446 & 2.095 & -1.133 & -0.772 & 2.977 & 1.067 \\
& w/o FSE
& 0.595 & 2.067 & -1.225 & -1.526 & 2.810 & 1.226 \\
& w/o RRS
& -0.334 & 2.110 & -1.006 & -1.885 & 2.918 & 1.600 \\
& w/o DGR
& 0.440 & 1.069 & -1.145 & -2.143 & 3.011 & 1.049 \\
& w/o IC/SHAP
& -0.027 & 2.409 & -1.460 & -1.590 & 2.922 & 1.200 \\
\midrule

\multirow{5}{*}{MDD$\downarrow$}
& Full GIFT
& 10.04 & 6.74 & 21.23 & 13.70 & 10.05 & 8.89 \\
& w/o FSE
& 9.34 & 6.24 & 26.74 & 15.12 & 11.26 & 9.75 \\
& w/o RRS
& 14.22 & 7.83 & 21.91 & 24.98 & 12.31 & 7.75 \\
& w/o DGR
& 10.35 & 8.30 & 21.86 & 25.62 & 9.90 & 10.02 \\
& w/o IC/SHAP
& 13.41 & 7.16 & 26.51 & 20.30 & 13.97 & 10.84 \\
\midrule

\multirow{5}{*}{Calmar$\uparrow$}
& Full GIFT
& 0.698 & 4.263 & -1.512 & -1.432 & 4.625 & 1.649 \\
& w/o FSE
& 0.941 & 4.027 & -1.634 & -2.024 & 3.604 & 2.122 \\
& w/o RRS
& -0.490 & 3.833 & -1.430 & -2.154 & 5.234 & 3.674 \\
& w/o DGR
& 0.716 & 1.826 & -1.529 & -2.181 & 4.582 & 1.799 \\
& w/o IC/SHAP
& -0.040 & 4.456 & -1.905 & -1.947 & 3.579 & 2.081 \\
\midrule

\multirow{5}{*}{Return$\uparrow$}
& Full GIFT
& 2.41 & 12.16 & -13.61 & -9.17 & 20.40 & 5.80 \\
& w/o FSE
& 3.25 & 10.58 & -18.32 & -12.69 & 17.59 & 8.23 \\
& w/o RRS
& -3.83 & 12.75 & -13.54 & -21.59 & 29.05 & 11.71 \\
& w/o DGR
& 2.53 & 6.03 & -14.16 & -21.87 & 19.88 & 7.02 \\
& w/o IC/SHAP
& -1.11 & 13.63 & -20.47 & -16.29 & 21.82 & 8.90 \\
\bottomrule
\end{tabular}
}
\end{table}

\paragraph{\bfseries\itshape Observations.}
The detailed results in Table~\ref{tab:ablation_details_all} are consistent with the averaged changes in Table~\ref{tab:ablation_main}.
Removing DGR yields the largest average degradation on Sharpe, Sortino, Calmar, and Return, suggesting that iterative diagnostic feedback is important for selecting a useful state-reward interface.
Removing IC/SHAP produces the largest average increase in MDD, indicating that factor-level informativeness and critic-attribution feedback are relevant to risk control.
Removing RRS mainly affects downside-sensitive metrics, especially Sortino and MDD, which is consistent with its role in shaping risk-aware training feedback.
Removing FSE leads to milder but generally negative average changes, suggesting that factor-guided state enhancement contributes to the interface but works best when coupled with reward shaping and diagnostic refinement.

Overall, these results support the design of GIFT as an integrated state-reward interface pipeline rather than a set of independent add-ons.

\section{Robustness Studies}
\label{app:ablation_robustness}

\subsection{Multi-Seed Performance Robustness}
\label{app:multiseed_robustness}

To test whether the main PPO comparison depends on a favorable random seed, we repeat the controlled comparison between GIFT+PPO and Pure PPO with seeds 123, 789, and 1024.
Table~\ref{tab:main_multiseed_pure_ppo_baseline} reports the mean and standard deviation across seeds.
This experiment focuses on Pure PPO because it is the cleanest no-test-leakage comparison: neither GIFT nor Pure PPO baseline can access the second half of the test window before final evaluation.
GIFT wins 65 of 90 seed-level metric comparisons, corresponding to a 72.2\% win rate, and reaches at least 4/5 averaged metric wins in five of six windows.
The main exception is W6, where Pure PPO baseline has slightly higher average return, Sharpe, and Calmar, while GIFT keeps lower average MDD and a similar Sortino.
This exception indicates return-side metrics in rising or recovery windows remain seed-sensitive, even though the downside-risk behavior is more stable.

\begin{table*}[t]
\centering
\caption{
Multi-seed controlled comparison on the Technology portfolio.
Values are mean$\pm$std over seeds 123, 789, and 1024.
Return and MDD are reported in percentages.
}
\label{tab:main_multiseed_pure_ppo_baseline}

\scriptsize
\setlength{\tabcolsep}{2.0pt}
\renewcommand{\arraystretch}{1.18}
\setlength{\aboverulesep}{0.35pt}
\setlength{\belowrulesep}{0.35pt}

\begin{adjustbox}{width=1.03\textwidth,center}
\begin{tabular}{lll@{\hspace{6pt}}*{10}{c}cc}

\toprule
\textbf{Window}
& \textbf{Test period}
& \textbf{Regime}
& \multicolumn{2}{c}{\textbf{Ret. (\%) $\uparrow$}}
& \multicolumn{2}{c}{\textbf{Sharpe $\uparrow$}}
& \multicolumn{2}{c}{\textbf{Sortino $\uparrow$}}
& \multicolumn{2}{c}{\textbf{MDD (\%) $\downarrow$}}
& \multicolumn{2}{c}{\textbf{Calmar $\uparrow$}}
& \winhead{Metric Win}
& \winhead{Seed Win} \\
\cmidrule(lr){4-5}
\cmidrule(lr){6-7}
\cmidrule(lr){8-9}
\cmidrule(lr){10-11}
\cmidrule(lr){12-13}
&
&
& \GIFThead{GIFT} & \basehead{Pure PPO}
& \GIFThead{GIFT} & \basehead{Pure PPO}
& \GIFThead{GIFT} & \basehead{Pure PPO}
& \GIFThead{GIFT} & \basehead{Pure PPO}
& \GIFThead{GIFT} & \basehead{Pure PPO}
& \winhead{} & \winhead{} \\
\midrule

W1
& 2020.12--2021.06
& Recovery bull
& \(\mathbf{0.47{\pm}2.36}\) & \(-3.78{\pm}1.87\)
& \(\mathbf{0.189{\pm}0.320}\) & \(-0.359{\pm}0.187\)
& \(\mathbf{0.187{\pm}0.325}\) & \(-0.352{\pm}0.184\)
& \(\mathbf{11.27{\pm}2.09}\) & \(12.97{\pm}1.74\)
& \(\mathbf{0.292{\pm}0.494}\) & \(-0.551{\pm}0.286\)
& \winfive{5/5}
& \textbf{14/15} \\

W2
& 2021.07--2021.12
& Late bull
& \(\mathbf{13.32{\pm}1.31}\) & \(13.18{\pm}0.97\)
& \(\mathbf{2.275{\pm}0.123}\) & \(2.198{\pm}0.121\)
& \(\mathbf{2.286{\pm}0.113}\) & \(2.218{\pm}0.121\)
& \(\mathbf{7.21{\pm}1.31}\) & \(7.41{\pm}0.50\)
& \(\mathbf{4.399{\pm}0.584}\) & \(4.208{\pm}0.544\)
& \winfive{5/5}
& \textbf{11/15} \\

W3
& 2021.12--2022.06
& Rate-hike bear
& \(\mathbf{-16.48{\pm}4.04}\) & \(-25.71{\pm}8.74\)
& \(\mathbf{-1.364{\pm}0.175}\) & \(-1.689{\pm}0.555\)
& \(\mathbf{-1.245{\pm}0.141}\) & \(-1.450{\pm}0.441\)
& \(\mathbf{24.42{\pm}3.38}\) & \(32.60{\pm}5.20\)
& \(\mathbf{-1.603{\pm}0.208}\) & \(-1.954{\pm}0.542\)
& \winfive{5/5}
& \textbf{10/15} \\

W4
& 2022.07--2022.12
& Volatile bottoming
& \(\mathbf{-17.24{\pm}6.79}\) & \(-22.27{\pm}1.57\)
& \(\mathbf{-1.483{\pm}0.461}\) & \(-1.999{\pm}0.294\)
& \(\mathbf{-1.579{\pm}0.462}\) & \(-2.123{\pm}0.279\)
& \(\mathbf{20.85{\pm}7.46}\) & \(25.73{\pm}1.98\)
& \(\mathbf{-1.995{\pm}0.175}\) & \(-2.211{\pm}0.097\)
& \winfive{5/5}
& \textbf{13/15} \\

W5
& 2022.12--2023.06
& Banking / AI rally
& \(\mathbf{21.77{\pm}6.45}\) & \(20.07{\pm}6.36\)
& \(\mathbf{2.736{\pm}0.379}\) & \(2.593{\pm}0.576\)
& \(\mathbf{3.036{\pm}0.348}\) & \(2.854{\pm}0.520\)
& \(\mathbf{9.78{\pm}1.46}\) & \(10.21{\pm}1.10\)
& \(\mathbf{5.244{\pm}2.256}\) & \(4.597{\pm}1.775\)
& \winfive{5/5}
& \textbf{10/15} \\

W6
& 2023.07--2023.12
& V-shaped recovery
& \(7.58{\pm}2.63\) & \(\mathbf{8.59{\pm}2.89}\)
& \(1.196{\pm}0.186\) & \(\mathbf{1.207{\pm}0.401}\)
& \(\mathbf{1.205{\pm}0.331}\) & \(1.198{\pm}0.410\)
& \(\mathbf{9.98{\pm}0.31}\) & \(11.03{\pm}1.96\)
& \(1.902{\pm}0.580\) & \(\mathbf{2.057{\pm}0.830}\)
& \wintwo{2/5}
& 7/15 \\

\midrule
\textbf{Summary}
& \multicolumn{12}{c}{5/6 windows reach at least 4/5 metric wins.}
& \winhead{5/6}
& \textbf{65/90} \\
\bottomrule
\end{tabular}
\end{adjustbox}
\end{table*}

\subsection{Robustness against a Stronger PPO Baseline}
\label{app:additional_train_plus_adapt_ppo_baseline_robustness}

We also compare against Baseline 2, a stronger train-plus-test PPO baseline.
Unlike Pure PPO baseline, additional train-plus-adapt PPO baseline first trains on the historical training period, continues PPO training on the first half of the test interval, and then evaluates on the second half.
This is an additional robustness check; the main controlled comparison remains Pure PPO baseline because additional train-plus-adapt PPO baseline uses more PPO training data.
Table~\ref{tab:additional_train-plus-adapt_PPO_baseline_completed_record} shows that GIFT remains competitive against this stronger baseline.
GIFT reaches at least 4/5 metric wins in all six windows and wins 28/30 metric comparisons overall, reducing the concern that the improvement depends only on a particular PPO baseline configuration.

\begin{table}[htbp]
\centering
\scriptsize
\setlength{\tabcolsep}{3.0pt}
\renewcommand{\arraystretch}{1.12}
\caption{Full robustness run against Baseline 2 on the Technology sector portfolio (seed 123). Each entry reports GIFT+PPO/Baseline 2. Return and MDD are reported in percentage points; Sharpe, Sortino, and Calmar are raw scores. Baseline 2 is a train-plus-test PPO baseline with additional historical PPO training. Lower MDD is better; higher is better for the other metrics.}
\label{tab:additional_train-plus-adapt_PPO_baseline_completed_record}
\resizebox{\columnwidth}{!}{
\begin{tabular}{lccccc c}
\toprule
\textbf{Window}
& \textbf{Return}
& \textbf{Sharpe}
& \textbf{Sortino}
& \textbf{MDD}
& \textbf{Calmar}
& \textbf{Win} \\
\midrule
W1
& 0.57/$-$7.37
& 0.167/$-$0.715
& 0.170/$-$0.719
& 11.67/15.75
& 0.258/$-$1.018
& \textbf{5/5} \\

W2
& 11.37/13.48
& 2.183/2.172
& 2.192/2.168
& 5.53/7.14
& 4.862/4.445
& \textbf{4/5} \\

W3
& $-$17.13/$-$22.24
& $-$1.497/$-$1.542
& $-$1.335/$-$1.339
& 23.43/29.53
& $-$1.775/$-$1.847
& \textbf{5/5} \\

W4
& $-$8.91/$-$17.93
& $-$0.657/$-$1.639
& $-$0.700/$-$1.847
& 13.46/21.15
& $-$1.383/$-$2.091
& \textbf{5/5} \\

W5
& 19.21/17.77
& 2.889/2.203
& 3.196/2.212
& 9.00/11.44
& 4.860/3.620
& \textbf{5/5} \\

W6
& 5.63/5.20
& 0.996/0.980
& 0.961/0.944
& 10.67/9.96
& 1.354/1.339
& \textbf{4/5} \\
\midrule
\textbf{Summary}
& \multicolumn{5}{c}{All 6 windows reach at least 4/5 metric wins}
& \textbf{28/30} \\
\bottomrule
\end{tabular}
}
\end{table}

\section{Case Studies of Generated Designs}
\label{app:case_studies}

This section provides supporting evidence for the adaptive-design analysis in Section~\ref{sec:semantic_analysis}.
The main text summarizes the selected state-reward interface for each rolling window in Table~\ref{tab:window_adaptive_design}.
Here we inspect the generated design archive at a finer granularity through operation-frequency statistics, a code-to-semantics example, and a refinement trajectory.
These analyses are descriptive: they show how GIFT composes financial state and reward operations, but they do not imply that the LLM discovers latent market regimes.

\subsection{Operation Frequencies and Generated Design Patterns}
\label{app:operation_frequency}

\begin{figure}[t]
    \centering
    \includegraphics[width=\columnwidth]{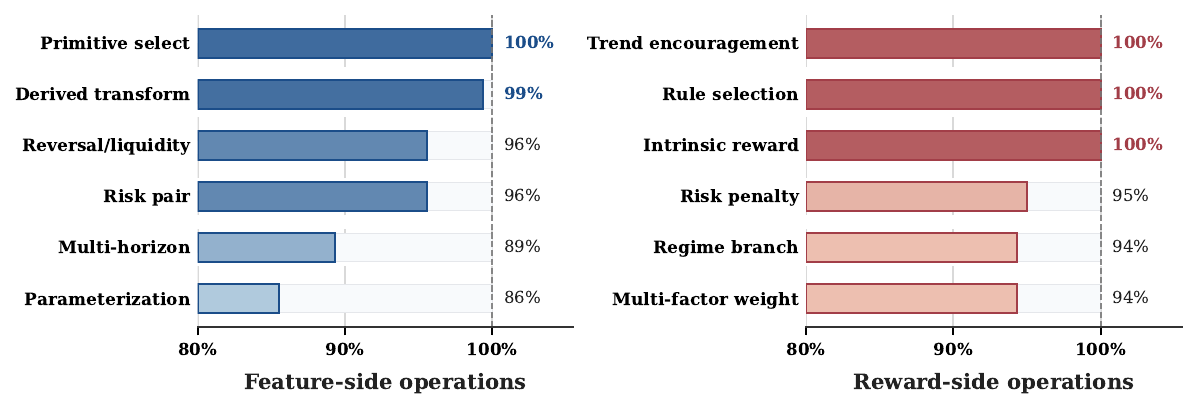}
    \caption{
    Operation-frequency analysis of parsed generated designs.
    Bars show the fraction of designs containing each feature-side or reward-side operation; darker colors indicate higher frequency, and categories are not mutually exclusive.
    }
    \label{fig:app_llm_operation_frequency}
\end{figure}

Figure~\ref{fig:app_llm_operation_frequency} summarizes operation tags parsed from generated designs. 
GIFT often combines primitive factors with transformations, return--risk terms, reversion or liquidity supplements, multi-horizon features, and explicit parameterization on the state side, while reward designs commonly include intrinsic rewards, rule selection, risk-threshold penalties, multi-factor weighting, and regime-conditioned branches. 
These patterns indicate that GIFT explores compositional state--reward interfaces rather than isolated indicators. 
They also broadly follow window-level market descriptions: recovery or rebound-like windows tend to use risk-threshold structures, whereas late-bull, rate-hike, and stress/rally windows more often combine trend, risk, reversion, and turnover terms. 
The generated interfaces therefore encode recognizable financial trade-offs while leaving final portfolio actions to PPO.

\subsection{W3 Code-to-Semantics Example}
\label{app:w3_code_semantics}

To make the semantic labels in Table~\ref{tab:window_adaptive_design} inspectable, we analyze the design selected at iteration 3 in W3.
This window corresponds to the rate-hike and geopolitical-shock bear market, and the selected design is summarized as ``mean reversion + state-reward coupling.''
The generated state function appends a mean-reversion channel, while the intrinsic reward includes a matching mean-reversion alignment bonus.
Thus, the same financial signal enters both the PPO observation interface and the reward feedback used during training.

\subsection{W3 Refinement Trajectory}
\label{app:w3_refinement_trajectory}
\begin{table}[H]
\centering
\scriptsize
\setlength{\tabcolsep}{3.8pt}
\renewcommand{\arraystretch}{1.18}
\caption{
Representative semantic evolution of generated designs in W3, the rate-hike and geopolitical-shock bear window.
}
\label{tab:app_feature_evolution}
\resizebox{\columnwidth}{!}{
\begin{tabular}{c p{0.38\columnwidth} c c p{0.30\columnwidth}}
\toprule
\textbf{Iteration}
& \textbf{Added Design Components}
& \textbf{Sharpe}
& \textbf{MDD}
& \textbf{Semantic Change} \\
\midrule
1
& Momentum and realized volatility
& 1.313
& 40.8\%
& Basic trend--risk frame \\

2
& Downside risk, multi-horizon momentum, and turnover
& 1.361
& 36.1\%
& Multi-factor expansion with liquidity \\

3
& Mean reversion and momentum-alignment reward rule
& \textbf{1.388}
& 45.5\%
& Reversion supplement with return-risk trade-off \\

4--5
& Z-score signal and stronger volatility/downside-risk penalties
& 1.22--1.24
& 39.9\%
& Conservative risk refinement \\

\bottomrule
\end{tabular}
}
\end{table}

Table~\ref{tab:app_feature_evolution} summarizes the W3 refinement trajectory. 
Across rounds, GIFT expands from momentum and volatility features to downside risk, multi-horizon momentum, turnover, mean reversion, and momentum-alignment rewards. 
The third-round design achieves the best validation Sharpe, while later, more heavily normalized or risk-penalized variants do not improve performance. 
This shows that GIFT selects the best validated executable interface, rather than the latest or most complex design.

\end{document}